%% file: main.tex
\ifdefined\pdftexversion\pdfoutput=1\fi %
\documentclass{article}
\PassOptionsToPackage{numbers,compress}{natbib}
\newif\ifarxiv
\input{edition}
\usepackage[preprint]{neurips_2026}

\usepackage[utf8]{inputenc}
\usepackage[T1]{fontenc}
\usepackage{hyperref}
\usepackage{url}
\usepackage{booktabs}
\usepackage{placeins} %
\usepackage{amsmath,amssymb,amsfonts}
\usepackage{graphicx}
\usepackage{nicefrac}
\usepackage{microtype}
\usepackage{xcolor}
\usepackage{listings}
\usepackage{etoolbox}
\makeatletter\AtBeginDocument{\@ifpackageloaded{lineno}{%
  \AtBeginEnvironment{lstlisting}{\par\nolinenumbers}\AfterEndEnvironment{lstlisting}{\par\linenumbers}}{}}\makeatother
\input{tables/numbers}
\newcommand{\CodeRepoParen}{(to be released)}
\newcommand{\SuppRef}{the code repository}
\input{tables/e2numbers}

\input{tables/e3numbers}

\input{tables/ablnumbers}

\input{tables/contamnumbers}
\input{tables/auxnumbers}

\input{tables/extnumbers}

\input{tables/benchnumbers}
\input{tables/holmmcnumbers} %
\ifdefined\ReproModels\else\input{tables/repronumbers}\fi %

\title{Grounding Language Models in Revealed Audience Behavior}
\hypersetup{pdftitle={Grounding Language Models in Revealed Audience Behavior}}

\input{authors}

\begin{document}
\maketitle

\begin{abstract}
Language models are increasingly used to simulate how an audience will respond to a headline or a post, and such simulators are validated against what people say.
We validate them against what people do.
We call a model grounded in revealed audience behavior when it is fine-tuned on records of what audiences actually clicked, liked or watched, and we ask not only whether asking a model falls short of grounding it, as prior work found, but why and where.
Why: the gap is a property of the training signal, not mainly of model size or prompting.
The same 8B model is right about half the time when asked which headline readers will click and \StrongestMOne\% of the time when grounded in the clicks; the strongest prompting we tried only matches a text classifier; a grounded 0.6B model beats every prompted frontier model in our runs; and whether the clicks are fitted as counts or as shares makes no detectable difference in a pre-registered test.
The asked 8B model also reproduces stated-preference cues, favoring question headlines that in fact lower clicks.
Where: across four behaviors, grounding wins where the text carries the signal (short-video likes, +20 points on 200 pre-registered videos), only ties an asked model given the reader's history (news clicks), and nothing text-only gets far from chance on watch time.
Code, adapters and run outputs will be released.
\end{abstract}

\section{Introduction}
\label{sec:intro}
Before publishing a headline, an advertisement or a post, a team would like to know how its audience will respond.
A cheap answer is to ask a language model, either directly or through a panel of simulated personas~\citep{argyle2023outofone,park2023generative,bui2025mop}.
The usual check compares such simulators with what people say, yet readers may dislike click-bait and still click it. We do not test people's own stated preferences. We compare prompted answers with clicks, as \citet{singh2025persuasion} and \citet{lu2026multiturn} do for likes and shopping actions, and we fixed the comparisons before the runs.

\begin{figure}[t]
\centering
\includegraphics[width=\linewidth]{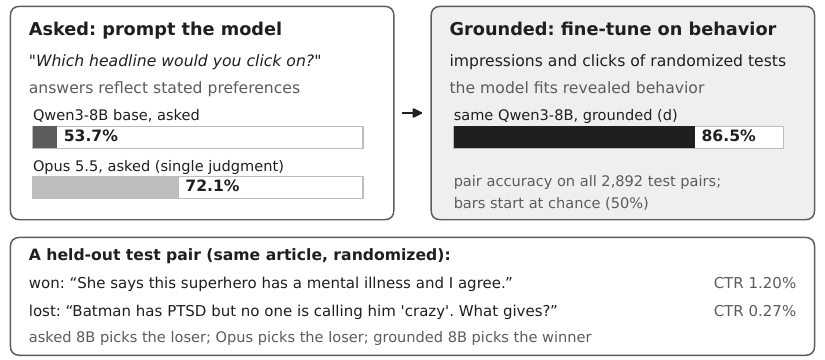}
\caption{Asked versus grounded. The same pretrained Qwen3-8B is right on \EtwoBaseAcc\% of the 2,892 held-out headline pairs when asked which headline it would click and on \StrongestMOne\% when grounded in impressions and clicks with option-distribution matching (d), the strongest baseline. Opus 5.5 asked once reaches \OpusFullMOne\%. Bottom: an illustrative test pair picked by a fixed rule, the shortest pair in which both asked models are wrong, (d) is right and only the loser has a question mark.}
\label{fig:overview}
\end{figure}

We ask whether the gap is real on behavioral data and whether it closes when we ground the same model.
Throughout, we call a model \emph{grounded} in revealed audience behavior when we fine-tune it on records of what audiences actually clicked, liked or watched, and \emph{asked} when we only prompt it for a prediction (Figure~\ref{fig:overview}).

All grounding techniques below are LoRA fine-tuning, and we name them only where they differ.
The Upworthy Research Archive~\citep{matias2021upworthy} records, for each randomized headline test, how often each variant was shown and clicked.
We hold one pretrained 8B model fixed and change only what we fit it to. We either ask it as a reader or ground it by regressing click rates, classifying pairs, matching the share of clicks over a test's variants (the pair-level objective), or fitting the impression and click counts with a binomial likelihood and one appeal term per experiment, so that a headline gets no credit for the article it introduces (the count-level objective, Section~\ref{sec:method}).

Claim 1 rests on the main comparison, fixed before the runs. Claim 2 adds a post hoc test (exploratory, not pre-specified).
\begin{enumerate}
\item \textbf{Accuracy.} \RegText{We test whether the count-level objective predicts the winner of held-out headline pairs better than the strongest supervised baseline (Section~\ref{sec:res-obj}).}%
{Every grounded objective predicts the winner of held-out headline pairs better than every untuned asked model, and the count-level objective beats the strongest supervised baseline in the main comparison (Section~\ref{sec:res-obj}).}%
{Every objective that grounds the 8B in clicks predicts the winner of held-out headline pairs better than every untuned asked model. The count-level objective does not beat the strongest of them, the pair-level objective, in the main comparison: we can exclude gains above \GateCIHi{} points but not smaller ones (Section~\ref{sec:res-obj}).}
\item \textbf{Asked versus revealed.} The answers of prompted models depart from revealed clicks, which shows up as feature effects with the wrong sign or size (Section~\ref{sec:res-e2}).
\end{enumerate}
We fixed the evaluation plan before the experiments and dated every later change in it.
We tune baselines on a development period inside the training data and select the strongest one on the test set, which can only make our comparison harder.

\paragraph{Contributions.} Unlike PersuasionBench~\citep{singh2025persuasion}, our headline data are randomized tests. Unlike it and LOLA~\citep{ye2024lola}, we fix the main comparisons before the runs and diagnose asked answers feature by feature.
(i) \textbf{Benchmark.} An asked-versus-grounded benchmark over four datasets, Upworthy, MIND~\citep{wu2020mind}, KuaiRand and MicroLens. On randomized headline tests one evaluator scores prompted judges, persona mixtures, an asked pretrained model, classifiers and grounded LoRA models.
(ii) \textbf{Method comparison.} We ground one pretrained 8B in clicks under four objectives: regression, pairwise classification, pair-level option-distribution matching and a count-level binomial objective on aggregate impressions and clicks. All four beat every untuned asked model, \RegText{and we report the count-level versus pair-level comparison when scored.}{and we report the count-level versus pair-level comparison as met.}{and fitting counts and fitting click shares reach pair accuracies within a point of each other in the main comparison, fixed before the runs.}
(iii) \textbf{Beyond headlines.} We run the same comparison \MindContribText{}. These observational logs mix creator, period and recommender exposure, a limit we state.
(iv) A script that regenerates every table from committed run outputs.

\input{related_work}
\input{method}
\input{results}

\section{Discussion and Limitations}
\label{sec:discussion}
\paragraph{Why and where.} In our runs the gap between asked and grounded models follows the training signal, not mainly model size or prompting, and it holds where the text carries the behavior, in headline clicks and short-video likes, but not in news clicks given reader history or in watch time.
\paragraph{Objective.} \RegText{We have not scored the main comparison yet.}%
{The count-level objective meets the pre-specified criterion: its pair accuracy on the clearly separated test pairs exceeds \StrongestBaseline{} by \GateDiff{} points and the paired bootstrap interval [\GateDiffCI] excludes zero. The margin rests on one seed per configuration and on more development runs for it than for the baselines (see Variance and power below), so we read it as evidence that the counts carry signal the pair-level objective misses, not as a measured size of that signal.}%
{On the clearly separated test pairs the two objectives are within a point of each other: the count-level objective differs from \StrongestBaseline{} by \GateDiff{} points [\GateDiffCI]. CTR regression trails both. A probe on the frozen 8B already closes part of the gap between (d) and (a), so the lead over the sentence classifier combines representation and objective.}
\paragraph{Scope of the data.} All experiments come from one English-language publisher in one period, 2013 to 2015, whose audience arrived mostly through social feeds. The base model may have seen this public archive, which published only winners. An exploratory verbatim-recall check in Appendix~\ref{app:contam} found that the model recalls winners no more often than losers. We did not test recall of click counts, and a LoRA could still read memory of the archive that verbatim recall does not show.
\paragraph{Aggregate data.} As Appendix~\ref{app:obslimits} details, we observe no reader, so the grounded models describe how an audience responded in aggregate, not how any person did. We do not model repeated exposure of the same reader, or overdispersion.
\paragraph{Variance and power.} \SeedText{We train each LoRA configuration once, with seed 0}{The main comparison uses seed 0. Retrained with seeds 1 and 2, the count-level model differs from \StrongestBaseline{} by \SeedDiff{} points on the three-seed means, with an interval that includes zero and is too wide to show equivalence}. Our comparisons cannot resolve differences below the minimum detectable difference of \AuxMdeProxy{} points. \ApiAgreeText{}

\bibliographystyle{plainnat}
\bibliography{refs}

\appendix
\input{appendix}

\end{document}

%% file: edition.tex
\arxivtrue

%% file: tables/numbers.tex
\newcommand{\StrongestBaseline}{(d)}

\newcommand{\StrongestMOne}{86.5}

\newcommand{\RunnerUp}{(e)}
\newcommand{\RunnerUpDiff}{+1.1}
\newcommand{\RunnerUpDiffCI}{$-$0.1, +2.2}
\newcommand{\MTwoRandom}{25.1}
\newcommand{\OpusFullMOne}{72.1}

\newcommand{\GateDiff}{+0.9}
\newcommand{\GateDiffCI}{$-$0.2, +2.0}
\newcommand{\GateCIHi}{2.0}
\newcommand{\SeedGroupMOne}{87.1}
\newcommand{\SeedGroupMOneCI}{85.9, 88.2}
\newcommand{\SeedDiff}{+0.4}
\newcommand{\SeedDiffCI}{$-$0.4, +1.3}

\newcommand{\SeedVarCI}{$-$0.6, +1.4}
\newcommand{\AskStrongMOne}{73.7}

\newcommand{\AskStrongMOneCI}{70.2--77.2}
\newcommand{\AskStrongDiffA}{+0.3}
\newcommand{\AskStrongDiffACI}{$-$4.3, +5.0}
\newcommand{\SizeSmallMOne}{83.2}
\newcommand{\SizeSmallMOneCI}{81.9--84.6}
\newcommand{\SizeSmallMOneSix}{81.0}
\newcommand{\SizeSmallDiffOpus}{+11.1}
\newcommand{\SizeSmallDiffOpusCI}{+9.2, +13.1}
\newcommand{\SizeSmallDiffStrong}{+7.3}
\newcommand{\SizeSmallDiffStrongCI}{+3.2, +11.5}
\newcommand{\SizeSmallSeedN}{3}
\newcommand{\SizeSmallSeedMin}{82.5}
\newcommand{\SizeSmallSeedMax}{83.2}
\newcommand{\SizeSmallSeedLoMin}{+8.5}
\newcommand{\SizeCountFourEight}{+0.4}
\newcommand{\SizeCountFourEightCI}{$-$0.6, +1.5}
\newcommand{\SizePairFourEight}{+1.5}
\newcommand{\SizePairFourEightCI}{+0.2, +2.7}
\newcommand{\LlamaMOne}{88.4}
\newcommand{\LlamaMOneCI}{87.3--89.6}
\newcommand{\LlamaDiffQwen}{+0.97}
\newcommand{\LlamaDiffQwenCI}{$-$0.03, +1.97}
\newcommand{\SizeSmallSubN}{14}
\newcommand{\SizeSmallSubLoN}{14}
\newcommand{\SizeSmallSubMinDiff}{+8.4}
\newcommand{\OtherFamMOne}{84.9}
\newcommand{\OtherFamDiffOpus}{+12.8}
\newcommand{\OtherFamDiffOpusCI}{+10.8, +14.7}
\newcommand{\RegText}[3]{#3}
\newcommand{\SeedText}[2]{#2}

%% file: tables/e2numbers.tex
\newcommand{\EtwoBaseAcc}{53.7}
\newcommand{\EtwoFtAcc}{85.8}
\newcommand{\EtwoOpusAcc}{72.1}
\newcommand{\EtwoQActual}{$-$0.174}
\newcommand{\EtwoQBase}{+0.075}
\newcommand{\EtwoQOpus}{$-$0.072}
\newcommand{\EtwoQFt}{$-$0.165}

\newcommand{\EtwoBaseTwoQReaderAcc}{53.67}
\newcommand{\EtwoBaseTwoQJudgeAcc}{56.95}
\newcommand{\EtwoBaseTwoQJudgeQ}{+0.101}
\newcommand{\EtwoBaseTwoQReaderQ}{+0.075}
\newcommand{\EtwoBaseTwoQActualQ}{$-$0.174}
\newcommand{\EtwoBaseTwoQAccDiff}{+3.28}
\newcommand{\EtwoBaseTwoQAccCI}{+1.42, +5.12}
\newcommand{\EtwoBaseTwoQReaderNeg}{$-$0.054}
\newcommand{\EtwoBaseTwoQJudgeNeg}{+0.030}
\newcommand{\EtwoBaseTwoQActualNeg}{+0.035}
\newcommand{\EtwoBaseTwoQNegDiff}{$-$0.084}
\newcommand{\EtwoBaseTwoQNegCI}{$-$0.121, $-$0.046}

\newcommand{\EtwoFullHolmAQwenReader}{1}
\newcommand{\EtwoFullHolmBQwen}{2}
\newcommand{\EtwoFullN}{2,292}
\newcommand{\EtwoFullNSix}{600}
\newcommand{\EtwoFullQActual}{$-$0.185}
\newcommand{\EtwoFullQMin}{$-$0.061}
\newcommand{\EtwoFullQMax}{+0.061}
\newcommand{\EtwoFullHolmAN}{11}
\newcommand{\EtwoFullHolmAM}{42}
\newcommand{\EtwoFullHolmBN}{5}
\newcommand{\EtwoFullHolmBM}{21}
\newcommand{\EtwoFullDiffMin}{$-$10.3}
\newcommand{\EtwoFullDiffMax}{$-$2.5}
\newcommand{\EtwoFullQwenReaderAgree}{26.7}
\newcommand{\EtwoFullQwenLogitCtr}{59.2}
\newcommand{\EtwoFullQwenLogitReader}{54.0}

\newcommand{\EtwoGAskedN}{4}
\newcommand{\EtwoGAskedM}{14}

\newcommand{\EtwoGOpusQ}{+0.072}
\newcommand{\EtwoGOpusQCI}{+0.034, +0.112}
\newcommand{\EtwoGTenN}{9}
\newcommand{\EtwoGTenM}{42}
\newcommand{\EtwoGTenQN}{4}

%% file: tables/e3numbers.tex
\newcommand{\EthreeSigN}{2}

\newcommand{\EthreeNumDiff}{$-$2.1}

\newcommand{\EthreeTimeDiff}{$-$3.3}

\newcommand{\EthreeSizePart}{$-$1.9}

\newcommand{\EthreeGapLessMin}{1.3}
\newcommand{\EthreeGapLessMax}{9.2}
\newcommand{\EthreeGapMoreMin}{1.2}
\newcommand{\EthreeGapMoreMax}{4.5}

%% file: tables/ablnumbers.tex
\newcommand{\AblK}{1}

\newcommand{\AblMainLoss}{0.8}
\newcommand{\AblSharedLoss}{11.6}
\newcommand{\AblMainMOne}{87.4}

\newcommand{\AblWinlossDiff}{$-$1.9}
\newcommand{\AblWinlossDiffCI}{$-$3.0, $-$0.9}

\newcommand{\AblSharedDiff}{$-$34.4}
\newcommand{\AblSharedDiffCI}{$-$36.5, $-$32.2}

%% file: tables/contamnumbers.tex
\newcommand{\ContamNTests}{300}
\newcommand{\ContamNItems}{600}
\newcommand{\ContamBaseExact}{0.17}
\newcommand{\ContamBaseExactCI}{0.00, 0.93}
\newcommand{\ContamDExact}{0.00}
\newcommand{\ContamDExactCI}{0.00, 0.61}
\newcommand{\ContamBaseExactWL}{$-$0.33}
\newcommand{\ContamBaseExactWLCI}{$-$1.00, 0.00}

\newcommand{\ContamBasePrefix}{4.6}

\newcommand{\ContamBasePrefixWL}{$-$1.2}
\newcommand{\ContamBasePrefixWLCI}{$-$2.8, +0.3}
\newcommand{\ContamDPrefixWL}{$-$0.9}
\newcommand{\ContamDPrefixWLCI}{$-$2.4, +0.5}
\newcommand{\ContamBaseFirst}{27.8}

\newcommand{\ContamGroupExact}{0.17}
\newcommand{\ContamGroupExactCI}{0.00, 0.93}
\newcommand{\ContamGroupPrefixWL}{$-$1.5}
\newcommand{\ContamGroupPrefixWLCI}{$-$3.3, +0.1}
\newcommand{\ContamGroupFirstWL}{$-$1.3}
\newcommand{\ContamGroupFirstWLCI}{$-$8.7, +6.3}
\newcommand{\ContamGroupMinusBasePrefix}{+0.3}
\newcommand{\ContamGroupMinusBasePrefixCI}{0.0, +0.6}
\newcommand{\ContamGroupMinusBaseFirst}{+0.8}
\newcommand{\ContamGroupMinusBaseFirstCI}{$-$0.8, +2.7}

%% file: tables/auxnumbers.tex
\newcommand{\AuxMdeProxy}{1.65}

\newcommand{\AuxCtrMedian}{1.0}
\newcommand{\AuxCtrHigh}{3.1}

\newcommand{\AuxLoraMdeQuestion}{4.0}
\newcommand{\AuxLoraMdeNumber}{4.2}
\newcommand{\AuxLoraMdeQuote}{7.3}
\newcommand{\AuxLoraMdeFirst}{4.2}
\newcommand{\AuxLoraMdeTime}{4.6}

\newcommand{\AuxAllDiff}{+0.85}
\newcommand{\AuxAllDiffCI}{+0.42, +1.28}

\newcommand{\AuxNPairs}{2,892}

\newcommand{\AuxSampleRule}{\texttt{pairs\_test.sample(600, random\_state=1)}}
\newcommand{\AuxAllNPairs}{38,931}

\newcommand{\AuxPrMOne}{81.3}

\newcommand{\AuxPrAMOne}{74.7}
\newcommand{\AuxPrDMOne}{86.5}
\newcommand{\AuxPrDiffA}{+6.6}

\newcommand{\AuxMdeSixLo}{3.7}
\newcommand{\AuxMdeSixHi}{5.4}

\newcommand{\AuxDcQuarterN}{1,884}
\newcommand{\AuxDcHalfN}{3,768}
\newcommand{\AuxDcFullN}{7,537}
\newcommand{\AuxDcQuarter}{77.8}
\newcommand{\AuxDcQuarterCI}{76.2, 79.3}
\newcommand{\AuxDcHalf}{84.5}
\newcommand{\AuxDcHalfCI}{83.2, 85.8}
\newcommand{\AuxDcFull}{86.5}
\newcommand{\AuxDcQuarterDiff}{$-$8.7}
\newcommand{\AuxDcQuarterDiffCI}{$-$10.2, $-$7.3}
\newcommand{\AuxDcHalfDiff}{$-$2.0}
\newcommand{\AuxDcHalfDiffCI}{$-$3.0, $-$1.0}
\newcommand{\AuxDcAQuarter}{73.6}
\newcommand{\AuxDcAHalf}{74.4}
\newcommand{\AuxDcAFull}{74.7}

\newcommand{\AuxAllPairsText}{Post hoc (rule written after the run), the gap on all \AuxAllNPairs{} within-experiment pairs is \AuxAllDiff{} [\AuxAllDiffCI]; the registered comparison is unchanged, and this result is only a candidate primary hypothesis for a new study.}

\newcommand{\AuxLoraMdeLead}{Using the pair-bootstrap standard error of the drop gap itself on each held-out format's test pairs (the same resamples as its interval)}

\newcommand{\AuxDcText}{\paragraph{Training-set size of (d) (exploratory).} Retraining (d) with its registered settings on a random quarter (\AuxDcQuarterN{} units) and half (\AuxDcHalfN{}) of the \AuxDcFullN{} training units, one draw each, gives \AuxDcQuarter\% [\AuxDcQuarterCI] and \AuxDcHalf\% [\AuxDcHalfCI] pair accuracy on all test pairs against \AuxDcFull\% with all units (\AuxDcQuarterDiff{} [\AuxDcQuarterDiffCI] and \AuxDcHalfDiff{} [\AuxDcHalfDiffCI] points). With the same shares of training pairs, (a) scores \AuxDcAQuarter\%, \AuxDcAHalf\% and \AuxDcAFull\% (mean of ten draws), so (d) stays ahead of (a) even with a quarter of the data. This curve was not used for the objective comparison or any claim.}

%% file: tables/extnumbers.tex
\newcommand{\MindContribText}{on a second click log (MIND news) and on a different behavior, watch time (KuaiRand), both with audience-conditioned variants, and on likes per view of short videos (MicroLens)}
\newcommand{\KwAccText}{with exposure removed by design (same viewer, day and length bin, random insertion), every method that ignores the viewer stays near chance (53.6--56.2\% on 3,000 pairs, including the 8B model trained on watch logs at 56.2\%; asked Sonnet 53.5\% and Haiku 52.2\% on a 1,000-pair subset), so the registered asked-versus-grounded test is not met (Holm) and cannot be decided with text representations of this data}
\newcommand{\KwAskedCondText}{on the same 1,000 pairs the viewer's recent history lifts the asked Sonnet judge by +5.85 [+2.96, +8.87] points and the 8B trained with it by +0.20 [$-$2.90, +3.29] (+0.53 on all 3,000 pairs), a paired difference of +5.65 [+1.70, +9.74] (post hoc, user-level bootstrap); the 8B result alone cannot tell a training setup that did not learn to use the history input from a history text that carries little watch-time signal for this model, although the same text lifts asked Sonnet}

\newcommand{\MindAppText}{Each asked MIND prompt was sent twice, the second time with the candidate order reversed, and an arm's top-1 hit rate in Table~\ref{tab:extfull} is the mean over the two orders, which cancels a constant preference for a position; an answer that is not a candidate letter counts as uniform over the candidates. Over the 1,000 impressions the two orders pick the same article in 66.9--77.1\% of impressions for Sonnet and in 33.9--55.2\% for Haiku (an order without a readable answer counts as a disagreement). Haiku gives no readable answer to 6.4--32.3\% of an arm's prompts (the higher of the two orders), Sonnet to none. Haiku's picks thus often change with the order or go unanswered, and the asked comparison rests on Sonnet. Asked instead which article of the list had the highest click rate, Sonnet reaches 30.7\% without and 30.4\% with the reader's history ($-$0.3 points [$-$2.5, +1.8]), so the history helps Sonnet only when it is asked about this reader. On the same 1,000 impressions the history lifts Sonnet asked as this reader by +9.35 [+6.04, +12.58] points and the grounded 8B by +9.20 [+5.50, +12.90], a paired difference of +0.15 [$-$4.11, +4.35] (post hoc, user-level bootstrap).}
\newcommand{\ExtBodyText}{On KuaiRand watch time the registered asked-versus-grounded test is not met (Holm); on the same 1,000 pairs, asking with the viewer's history gives 59.4\% vs.\ at most 56.1\% for viewer-blind methods (untested). MIND's 8B lead over asked Sonnet is not confirmed on 3,000 new impressions; MicroLens frame text adds +7.3 and +7.0 points to the 8B and the classifier (Holm), and its 8B lead over asked Sonnet is confirmed on 200 new videos (Holm; with API answers, frames only; lengths matched: title +8.1 [$-$4.8, +21.0], frames +15.3 [+3.2, +26.6]; Table~\ref{tab:extfull}).}
\newcommand{\MlFramesAppText}{For the ``+ frames'' rows of Table~\ref{tab:extfull}, ffmpeg took four still frames from each video at 12.5\%, 37.5\%, 62.5\% and 87.5\% of its length, and Claude Haiku 5.5 wrote one English paragraph about them under a fixed, registered prompt. The prompt asks for people, animals, objects, setting, actions and framing, forbids judging popularity, quality, appeal or viewers' emotion, and forbids transcribing on-screen text. Each of the 4,747 videos was described once, and every method receives the same text. 677 descriptions came through the Claude Code command-line interface and, after its session limit was reached, 4,070 through OpenRouter (\texttt{anthropic/claude-haiku-5.5}, temperature 0, US\$1.16 in total). The access path is balanced across the label (677: 344 lower and 333 upper; 4,070: 2,030 and 2,040), but it changes length: the prompt asks for 60--100 words, while the median description has 114 words through the command-line interface and 140 through OpenRouter. The frame gains may therefore include a small path effect, which we do not separate.}
\newcommand{\AudienceText}{Who the reader is matters: on the same 1,000 items, history lifts Sonnet asked as this reader by +9.35 points on MIND and +5.85 on KuaiRand, and the grounded 8B by +9.20 on MIND (+9.26 on 20,000 impressions, Holm) but +0.20 on KuaiRand (intervals in App.~\ref{app:behaviors}); it does not lift Sonnet asked for the list's click rate ($-$0.30 [$-$2.51, +1.85]).}
\newcommand{\ApiPathAppText}{The MicroLens and MIND asked cells were answered through that client under the default system prompt of Claude Code, which includes its tool definitions (about 36,700 input tokens per call), at a sampling temperature that cannot be set. As registered on 17 Nov 2026, we re-ran five of these cells through the API (Sonnet 5.5, temperature 0, no added context): the four MicroLens cells and the MIND confirmation cell; the 1,000-impression MIND sample keeps only its command-line answers. We call a cell the same when the paired 95\% interval of API minus command-line accuracy includes 0; the first-sample frames cell differs (+5.4 points [+1.8, +8.9]), so the MicroLens asked values in the main text are API answers, with the command-line values beside them in the caption of Table~\ref{tab:extfull}. The registered MicroLens confirmation stays on the command-line answers (title +11.0, Holm $p=0.019$, confirmed; frames +20.0, confirmed); with the API answers the title lead is +9.0 with Holm $p=0.067$ (not confirmed) and the frames lead +20.0 (confirmed). The MIND confirmation cell is the same (top-1 $-$0.8 points [$-$1.7, +0.1]), so the MIND confirmation result does not depend on the access path; it is not confirmed on either path ($-$0.5 and +0.3 points). The two paths reach similar accuracy with different answers: they agree on 64--73\% of MicroLens videos and 79\% of MIND answers, so single asked answers should not be read as stable.}
\newcommand{\ApiAgreeText}{Asked answers also depend on the access path: the same model, asked the same items through the API and through the command-line client, agrees on 64--79\% of its answers.}
\newcommand{\MindAbsText}{On news clicks (MIND), the reader's history lifts an 8B model grounded in these clicks by +9.3 points [+8.5, +10.1] (Holm), but it is not significantly above asking Sonnet as this reader on 1,000 registered or 3,000 new impressions ($-$0.5 [$-$2.1, +1.2]).}

\newcommand{\KwMdeText}{Post hoc, with the standard error read from each paired interval, the minimum detectable difference (MDE) of the registered comparisons at 80\% power is 3.3--3.9 points (the 8B against the past long-view rate) and 5.2--6.2 points (against asked Sonnet on the 1,000-pair subset), from two-sided 5\% to the strictest Holm step; the intervals rule out an 8B lead above 4.9 and 6.0 points, so a smaller gain from training is neither shown nor excluded.}
\newcommand{\MindMdeText}{On the 3,000 new impressions the top-1 interval rules out an 8B lead above 1.2 points and a deficit above 2.1 points; at 80\% power and two-sided 5\% the MDE of the comparison is about 2.4 points (post hoc, standard error from the interval width).}
\newcommand{\MlMdeText}{Of the MicroLens confirmation checks that are not significant, at 80\% power and two-sided 5\% the title comparison with API answers (200 videos, +9.0 [+0.0, +18.5]) has an MDE of about 13.2 points and the title comparison on the 124 length-matched videos (+8.1 [$-$4.8, +21.0]) about 18.4 (post hoc, standard error from the interval width). Both intervals contain the first sample's title lead (+11.2 points, API answers), so these checks lack the power to confirm it rather than speaking against it.}
\newcommand{\MlLengthText}{Outside the registered family, the grounded 8B leads a log-length logistic model by +3.2 [$-$0.3, +6.9] points on the 942 first-sample videos and a length-plus-category model by +3.0 [$-$0.6, +6.7]; both intervals include zero, so on this sample length alone may explain most of the grounded lead. On the 621 length-matched videos, where the length model is at chance (49.3\%), the lead is +15.3 [+10.3, +20.3].}

\newcommand{\ExtShortText}{Outside headlines the grounded lead depends on the behavior and on knowing the reader. On news clicks (MIND), the reader's history lifts Sonnet asked as this reader as much as the grounded 8B (+9.35 and +9.20 points on the same items), and on new impressions the 8B is not confirmed above asking. On KuaiRand watch time the grounded 8B fails the main test against asking, and history barely moves it (+0.20 points). On short-video likes (MicroLens), frame text adds +7.3 points to the 8B after Holm correction, and on new videos its lead over asked Sonnet holds with frames and API answers (Table~\ref{tab:extfull}).}

%% file: tables/benchnumbers.tex
\newcommand{\BenchBodyText}{On the asked-versus-grounded benchmark's scale the grounded 8B leads the asked LLM by 2.9 to 31.0 points, the interval including 0 on two of four (post hoc, Table~\ref{tab:bench}).}

\newcommand{\BenchAppText}{\paragraph{Asked-versus-grounded benchmark.} Table~\ref{tab:bench} puts the asked and grounded cells of the four datasets on one scale, as SimBench~\citep{hu2026simbench} does for simulators. The gap is large on Upworthy (31.0 points) and MicroLens (22.3); on MIND (2.9) and KuaiRand (4.8) its interval includes 0 and both arms score at most 11.8 of 100. The rule was written before the computation (registration log, 18 November) but after all cell values were known, so this is a summary, not a test.}

%% file: tables/holmmcnumbers.tex
\newcommand{\HolmMcFamWithP}{9}
\newcommand{\HolmMcFamStable}{8}
\newcommand{\HolmMcETwoMin}{4}
\newcommand{\HolmMcETwoMax}{6}
\newcommand{\HolmMcETwoFlipN}{2}
\newcommand{\HolmMcETwoFlipZ}{1.1}
\newcommand{\HolmMcKse}{3}

%% file: tables/repronumbers.tex
\newcommand{\ReproModels}{\texttt{anthropic/claude-haiku-5.5}, \texttt{anthropic/claude-sonnet-5.5}, \texttt{anthropic/claude-opus-5.5}}
\newcommand{\ReproJudgeDate}{12 Nov 2026, 02:34}
\newcommand{\ReproMopDate}{12 Nov 2026, 02:30}
\newcommand{\ReproOpusFullDate}{12 Nov 2026, 12:50}

\newcommand{\ReproBaseAskN}{2,892}
\newcommand{\ReproBetaOne}{0.9}
\newcommand{\ReproBetaTwo}{0.999}
\newcommand{\ReproEps}{10^{-8}}

\newcommand{\ReproPrefixLen}{4}

\newcommand{\ReproTrainMinTotal}{4.0}
\newcommand{\ReproPrN}{149}
\newcommand{\ReproPrPzero}{3}
\newcommand{\ReproPrPone}{56}
\newcommand{\ReproPrPthree}{90}
\newcommand{\ReproPrRule}{13}
\newcommand{\ReproPrPoneRule}{11}
\newcommand{\ReproPrClaim}{8}
\newcommand{\ReproPrAux}{53}
\newcommand{\ReproPrPost}{54}
\newcommand{\ReproPrRecord}{21}
\newcommand{\ReproPrPthreePost}{44}
\newcommand{\ReproPrPthreeRecord}{5}
\newcommand{\ReproPrPthreeAux}{38}
\newcommand{\ReproPrPthreeClaim}{3}
\newcommand{\ReproPrWrittenAfter}{5}
\newcommand{\ReproPrMdeDate}{12 Nov 2026, 12:20}

\newcommand{\ReproCliOrAcc}{60.8}
\newcommand{\ReproCliAcc}{56.5}
\newcommand{\ReproCliDiff}{$-$4.3}
\newcommand{\ReproCliDiffCI}{$-$8.8, +0.0}
\newcommand{\ReproCliBatches}{60}
\newcommand{\ReproCliRefused}{5}
\newcommand{\ReproCliPreamble}{21}
\newcommand{\ReproCliPairs}{600}
\newcommand{\ReproCliVersion}{2.1.310}
\newcommand{\ReproCliRegOrAcc}{60.3}
\newcommand{\ReproCliRegAcc}{53.2}
\newcommand{\ReproCliRegDiff}{$-$7.2}
\newcommand{\ReproCliRegDiffCI}{$-$11.7, $-$2.7}
\newcommand{\ReproCliNoAns}{50}
\newcommand{\ReproCliOrNoAns}{5}

%% file: authors.tex
\author{Jiwoo Jung\thanks{Corresponding author.} \\ Sigmine \\ \texttt{jiwoochris@sigmine.ai} \And Daecheol An \\ Sigmine \\ \texttt{andaniel@sigmine.ai}}
\hypersetup{pdfauthor={Jiwoo Jung, Daecheol An}}

%% file: related_work.tex
\section{Related Work}
\label{sec:related}

\paragraph{LLMs as simulated people.}
A growing line of work asks large language models to stand in for human respondents:
conditioning on demographic backstories to reproduce survey marginals~\citep{argyle2023outofone},
replicating classic subject studies~\citep{aher2023turing},
measuring whose opinions model outputs reflect~\citep{santurkar2023opinions},
and populating interactive agent societies~\citep{park2023generative} or recommender sandboxes~\citep{zhang2023agent4rec}.
SimBench~\citep{hu2026simbench} unifies 20 such datasets and reports only modest group-level fidelity (40.80/100) for the best current models.
These evaluations score a simulator against what people \emph{say} in surveys or against stylised tasks, though even people's stated priorities say little about what their ratings reward~\citep{chen2026stated}.
Persona agents trail a TF-IDF classifier at predicting reactions to posts~\citep{bojic2026persona}, and simulated readers overstate headline-rewrite effects~\citep{feroz2026debias}.
Against logged behavior, LLMs drift towards a hyper-active ``positive average person''~\citep{chen2026omnibehavior}, and simulated customers voice half the resistance of real non-buyers~\citep{chen2026walkaway}.
A review of over 100 simulation studies finds that they often get the direction of effects right but are far less reliable about sizes and distributions, and discusses correcting the error with human data close to the target setting~\citep{graeber2026limits}.
Closest in design, PersuasionBench~\citep{singh2025persuasion} asks which of two same-brand tweets earned more likes: 10-shot GPT-4o is right on 60--62\% and a 13B model trained on like counts on 76--81\%.
ShopCART~\citep{lu2026multiturn} finds that a 7B model fine-tuned on click logs predicts shoppers' next actions better than prompted frontier models.
UserMirrorer~\citep{wei2026usermirrorer} trains a 3B simulator on recommender feedback but on MIND only matches a frontier model, and linear probes on hidden states recover group opinions that next-token answers miss~\citep{jahanparast2026opinions}.
Conversely, zero-shot role-play reproduces the correlation structure of personality scales~\citep{liu2026fivedims}.
We instead score models against what people \emph{did} in randomized field experiments. Unlike these comparisons, ours spans four datasets, compares two objectives that learn from those outcomes, and fixes the headline comparison and the new-item checks on news and short videos before the runs.

\paragraph{Fitting simulators to aggregate response distributions.}
Closest to our method are approaches that fit a model to group-level outcome distributions rather than to individual labels.
\citet{suh2025survey} fine-tune pretrained base models so that the first-token probabilities of answer letters match survey response proportions under a forward KL objective.
\citet{cao2025specializing} specialize LLMs to survey response distributions across countries, and
Centaur~\citep{binz2025centaur} and Chiron~\citep{ye2026chiron} fine-tune foundation models on trial-level data from many psychology and social-science experiments.
Mixture-of-Personas~\citep{bui2025mop} represents a population as a learned mixture of persona-and-exemplar prompts, without fine-tuning.
All of these supervise with \emph{stated} responses from groups whose identity they observe.
Our setting has neither: Upworthy experiments report only impressions and clicks per headline, with no reader attributes.
We fit these counts in two ways (Section~\ref{sec:method}): a count-level objective, the binomial likelihood of each headline's clicks given its impressions with a per-experiment article appeal term, and a pair-level objective in the style of \citet{suh2025survey}, which matches answer-letter probabilities to the click shares within one experiment.
Both use the same click data, and neither uses reader attributes.

\paragraph{Predicting content experiments from text.}
The Upworthy Research Archive~\citep{matias2021upworthy} releases tens of thousands of headline A/B tests run on a news site. In it, numerals lift click-through by 5.2\% after controls~\citep{peters2026numerals}.
LOLA~\citep{ye2024lola} uses it to predict which headline wins an experiment and reports that a LoRA-tuned 8B model and an embedding regressor are close,
at 46.86\% and 46.28\% top-1 on a random experiment split in arXiv v3. With a 0.6B LoRA model the embedding regressor leads, 42.79\% to 35.70\%~\citep{sirsat2026lora}.
\citet{maiorano2026personas} compare a ten-persona LLM panel with a single generic-reader prompt on Upworthy and find that the panel ranks headlines worse.
\citet{persson2026surrogacy} give conditions under which calibrated LLM outcomes identify a human treatment effect and report that raw LLM outputs recover 39\% of the Upworthy effect. On 183 Upworthy interventions, offline-eval effects track online ones only weakly~\citep{schultzberg2026trial}.
Two 2026 preprints simulate A/B outcomes with agents conditioned on behavioral profiles or data-driven personas and report 0.70--0.80 directional agreement with historical tests~\citep{hut2026srct,benomar2026personaab}, and \citet{hut2026srct} add a pre-period calibration step.
\citet{loghmani2025observational} warn that fine-tuning on observational engagement data learns confounded correlations.
We report pair accuracy and LOLA top-1 on randomized tests only, with a time split fixed in advance.
We review work on generating creatives for clicks in Appendix~\ref{app:morerelated}.

%% file: method.tex
\section{Data and Evaluation}
\label{sec:data}

\paragraph{Experiments.}
We use the exploratory and confirmatory releases of the Upworthy Research Archive~\citep{matias2021upworthy}: 128,217 rows from January 2013 to April 2015.
Each row reports, for one headline variant in a randomized test, the number of impressions $n$ and clicks $c$.
Following the archive's 2024 correction~\citep{matias2024correction}, we drop experiments created on or after 2013-06-25 and before 2014-01-11, which have a known randomization imbalance.
We keep rows with $n>0$ and a headline, without duplicates.

\paragraph{Units.}
A \emph{unit} $i$ is a set of headlines $h_{i1},\dots,h_{iJ_i}$ tested on the same article with the same image and lead paragraph, so that the headline is the only thing that varies.
We merge rows with identical text inside a unit by summing $n_{ij}$ and $c_{ij}$ and drop units with fewer than two distinct headlines.
All units of one experiment fall in the same split.

\paragraph{Pairs.}
We also use the pair data of our pilot run, which predates our evaluation plan. From each experiment it takes the most clearly separated pair, the one with the smallest $p$, among headlines that share image and lead paragraph and whose click rates differ under a two-proportion $z$-test at $p<0.05$, and flips the A/B order at random with seed 0.

\paragraph{Temporal split.}
Everything created before 2014-10-01 is training data and everything after is test data. This gives 4,928 training and 2,892 test pairs, and 7,537 training and 4,407 test units, the latter from 4,305 experiments. A unit has 4.28 headlines on average over both sets and at most 17.
Every test pair lies inside one test unit.
All tuning happens inside the training period: the 2,932 units created on or after 2014-07-01 form the development set, and we fit on the 4,605 earlier units during tuning.
We refit final models on all training units and report the main comparison whichever way it falls.
We used the test set only once per final model. We committed the evaluation rules, comparisons and baselines to a pre-registration file before the corresponding runs and logged every later change there with its date and reason.

\paragraph{Metrics.}
One evaluator scores all methods.
\begin{itemize}
\item \textbf{Pair accuracy (M1)}: the fraction of test pairs in which the method's score is higher for the headline with the higher observed click rate. Ties and missing answers count as errors. We report all 2,892 pairs and the 600-pair pilot subset, on which we ran the commercial LLM judges. Pair accuracy covers only the 2,892 of 4,305 test experiments, 67\%, that have such a pair. We score the commercial judges on pair accuracy only, to limit API cost.
\item \textbf{Top-1 hit rate (M2)}: the fraction of test units in which the highest-scoring headline has the highest observed click rate. Score ties count by their expected value, and chance is the mean of $1/J_i$.
\item \textbf{Log-likelihood gain (M3)} uses the held-out binomial likelihood. We map a method's headline score $s_{ij}$ to a click rate $r_{ij}=\sigma(\alpha_i+\beta s_{ij})$, fit the single scalar $\beta$ on training units and set the unit intercept $\alpha_i$ so that predicted clicks match the unit's total, $\sum_j n_{ij}\,\sigma(\alpha_i+\beta s_{ij})=\sum_j c_{ij}$. The within-unit ranking therefore never enters the fit of $\alpha_i$. We report the gain in log-likelihood over the null model in which all headlines of a unit share one click rate, in nats per 1,000 impressions.
\end{itemize}
Confidence intervals are percentile intervals from 10,000 bootstrap resamples of test experiments with seed 0, except 2,000 for the contamination check and the interval that also resamples seeds, 10,000 resamples of users for KuaiRand and 2,000 of users or videos for MIND and MicroLens. Differences between two methods use the same resamples, a paired bootstrap. We Holm-adjust families of tests, and Table~\ref{tab:holm} lists them all.
Since there is one pair per experiment, resampling experiments is the same as resampling pairs for pair accuracy.

\section{Method: Objectives That Learn from Revealed Behavior}
\label{sec:method}

We ground the same pretrained model in the same click records with two objectives that differ in what they read from a unit: the counts each headline received (count-level) or the click-rate shares of the headlines shown together (pair-level).

\paragraph{Count-level objective.}
One pretrained base language model with a LoRA adapter reads the prompt
\begin{quote}\small\ttfamily\raggedright\frenchspacing
Headline: \{h\}\textbackslash n Would this reader click this headline? Answer click or skip.\textbackslash n Answer:
\end{quote}
preceded by $P=4$ trainable prompt vectors of the model width $d$, and its click logit is the difference of the next-token logits for ``\texttt{\ click}'' and ``\texttt{\ skip}'',
\begin{equation}
\ell(h)=\operatorname{logit} p(h),\qquad p(h)=\frac{p(\texttt{click}\mid h)}{p(\texttt{click}\mid h)+p(\texttt{skip}\mid h)} .
\end{equation}
With an article appeal term $\alpha_i$ for unit $i$, the predicted click rate of headline $j$ is
\begin{equation}
\label{eq:mix}
r_{ij}=\sigma\!\big(\alpha_i+\ell(h_{ij})\big),
\end{equation}
so the article's overall appeal cannot be credited to the headline.
We maximize the binomial log-likelihood of the observed impressions and clicks,
\begin{equation}
\label{eq:ll}
\mathcal{L}=\sum_i\sum_j c_{ij}\log r_{ij}+(n_{ij}-c_{ij})\log(1-r_{ij}),
\end{equation}
over the LoRA weights, the prompt vectors and the free per-unit $\alpha_i$.
We use no reader attributes and no pairwise labels, only the counts each headline received.
We minimize the binomial deviance, the saturated minus the model log-likelihood, averaged over headlines in the batch, a constant shift and rescaling of $-\mathcal{L}$.
Units without clicks do not inform $\alpha_i$, and we skip them.

\paragraph{Profiling out the article appeal.}
Rather than learning one parameter per training unit, we profile $\alpha_i$ out: for the current $\ell$, held fixed without gradient, 25 safeguarded Newton steps on the unit's own likelihood find $\hat\alpha_i=\arg\max_{\alpha}\mathcal{L}_i(\alpha)$, starting from the logit of the unit's pooled click rate minus the mean logit.
Because $\partial\mathcal{L}_i/\partial\alpha_i=0$ at $\hat\alpha_i$, the gradient of the profiled objective with respect to the LoRA weights and prompt vectors equals that of the joint fit by the envelope theorem.

\paragraph{Inference.}
On a test unit we set $\alpha_i$ by matching total clicks, exactly as for the log-likelihood gain, so that the same rule applies to all methods. We then evaluate $r_{ij}$ in~\eqref{eq:mix} directly, without a fitted $\beta$.
The pair score is the difference of $\ell$ between the two headlines, which does not depend on $\alpha_i$.
We report the log-likelihood gain with an additional fitted $\beta$ and the model's own $\alpha_i$ estimate as auxiliary numbers.

\paragraph{Pair-level objective.}
Baseline (d) below fits the same base model to the same units through their click-rate shares: it shows all headlines of a unit in one prompt and matches the option-letter distribution to the shares with forward KL.\@
Click rates are small: on training units the impression-weighted median headline click rate is \AuxCtrMedian\% and the 95th percentile \AuxCtrHigh\%. At such rates profiling $\alpha_i$ out of~\eqref{eq:ll} leaves the multinomial likelihood of the unit's clicks under $\operatorname{softmax}_j(\ell(h_{ij})+\log n_{ij})$, which with this archive's nearly equal impressions is the objective of (d) weighted by the unit's clicks (Appendix~\ref{app:multinomial}).
The two objectives are thus close to a pointwise and a listwise parameterization of nearly the same likelihood. They differ in the weight of a unit, the exact binomial terms and whether the model scores a headline alone or next to its siblings.

\paragraph{Ablations.}
Two ablations, fixed in advance, each change one part of the count-level objective.
(i) \emph{Win/loss only} replaces the binomial likelihood by the mean pairwise cross-entropy $-\log\sigma(s_{ij}-s_{ij'})$ over within-unit pairs with $c_{ij}/n_{ij}>c_{ij'}/n_{ij'}$, where $s_{ij}=\ell(h_{ij})$.
(ii) \emph{No article appeal} replaces the per-unit $\alpha_i$ by one trainable $\alpha$ shared by all units, initialized at the pooled logit click rate.

\section{Baselines}
\label{sec:baselines}

All baselines use the same split and evaluator. Fine-tuned baselines share our base model.
\begin{itemize}
\item[(a)] \textbf{Embedding + logistic regression.} Logistic regression on the difference of the two headlines' all-mpnet-base-v2 sentence embeddings, fitted on the 4,928 training pairs. We choose the regularization $C$ among 10 values from $10^{-4}$ to $10^{4}$ by mean validation accuracy over five repetitions of shuffled stratified 5-fold cross-validation with seeds 0--4 and refit to a tolerance of $10^{-10}$. Top-1 hit rate and log-likelihood gain use the linear score of one embedding.
\item[(b)] \textbf{TF-IDF + logistic regression.} The same on word 1--2-gram TF-IDF vectors with sublinear term frequency and minimum document frequency 2, with $C$ chosen among 10 values by one 5-fold cross-validation on the training pairs.
\item[(c)] \textbf{LOLA-style regression}~\citep{ye2024lola}. With one headline in the prompt, the RMS-normalized last-token hidden state feeds a linear head that regresses the empirical logit click rate, $\log\frac{c+0.5}{n-c+0.5}$, with mean squared error. Unlike the original, which regresses the raw rate, we subtract the unit mean of the target so the head ignores article appeal. The raw-target variant is auxiliary.
\item[(d)] \textbf{Suh-style option distribution}~\citep{suh2025survey}. We list up to 8 headlines of a unit, a random subset when there are more, as options A, B, \dots\ in a fresh order each time we draw the unit, after ``Several headlines were tested on the same article. Which one did readers click the most?''. We fit the softmax over the first-token probabilities of the option letters to the unit's click-rate shares $\big(c_{ij}/n_{ij}\big)/\sum_{j'} c_{ij'}/n_{ij'}$ with forward KL.\@ At test time we score each unit in $J_i$ cyclic windows of $\min(J_i,8)$ consecutive headlines, so that every headline appears once in every shown position, and a headline's score is its mean centered log-probability. It sees the same click data as the count-level objective, as within-unit shares: it is the pair-level objective of Section~\ref{sec:method}.
\item[(e)] \textbf{Pairwise LoRA classifier.} ``Two headlines were tested on the same article. Which one got the higher click rate? A: \dots\ B: \dots'' with cross-entropy on the first token ``A''/``B''.\@ Training pairs are all within-unit pairs whose click rates differ significantly under the pair rule above, each in both orders. We antisymmetrize the pair score over the two orders. Headline scores for top-1 hit rate and log-likelihood gain are round-robin means against the other headlines of the unit.
\item[(f)] \textbf{Commercial LLM judges.} We run Claude Sonnet 5.5 and Opus 5.5 once on the 600-pair subset, and Opus 5.5 also on all 2,892 test pairs, with the pilot's prompt unchanged: ten pairs per request, each shown as X/Y in a pair-specific random order, at temperature 0.7.
\item[(g)] \textbf{Mixture-of-Personas prompting}~\citep{bui2025mop}. A fitted gate combines eight LLM-written personas with in-context examples. We run it with Haiku and Sonnet 5.5 on the 600-pair subset.
\end{itemize}
The \emph{strongest baseline} is the one among (a)--(e) with the highest pair accuracy on all 2,892 test pairs.
Selecting it on the test set favors the baselines, so it can only make our comparison harder.
As a reference line we also report (h), a logistic regression on seven surface features: second-person, first-person, question mark, digit, negation, length and emphasis. It is not a candidate for the strongest baseline (reference rows in the supplementary appendix to be released with the code{}).

\paragraph{Comparison.}
Our main comparison is the pair-accuracy difference between the count-level model and the strongest baseline on the 2,892 test pairs, with its 95\% paired bootstrap interval. We report top-1 hit rate and log-likelihood gain alongside.
Against (d) it compares pointwise with listwise scoring of nearly the same likelihood. All fine-tuned models are rank-16 LoRA adapters on a frozen Qwen3-8B-Base, with training details in Appendix~\ref{app:training}.

%% file: results.tex
\ifdefined\StrongestBaseline\else\input{tables/numbers}\fi
\ifdefined\EtwoBaseAcc\else\input{tables/e2numbers}\fi
\ifdefined\EthreeSigN\else\input{tables/e3numbers}\fi
\ifdefined\AblK\else\input{tables/ablnumbers}\fi
\ifdefined\UNineBodyText\else\input{tables/u9numbers}\fi
\ifdefined\BenchBodyText\else\input{tables/benchnumbers}\fi
\ifdefined\UThreeBodyText\else\input{tables/u3numbers}\fi
\section{Results}
\label{sec:results}
The results answer two questions: why asking falls short of grounding on the same headlines, through the baselines, the objective comparison and the feature diagnostic, and where grounding helps across behaviors.

\input{tables/stage1}
\subsection{Baselines}
\label{sec:res-stage1}
\begin{figure}[t]
\centering
\includegraphics[width=0.8\linewidth]{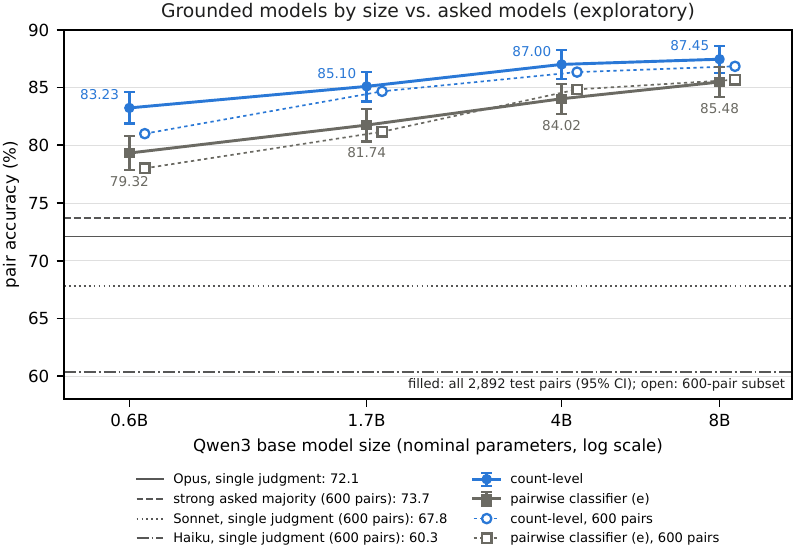}
\caption{Pair accuracy (M1) on the 2,892 test pairs against base-model size, with 95\% bootstrap intervals, exploratory. Blue circles: count-level model. Gray squares: pairwise classifier (e). Same settings at every size. Open dotted markers: the same runs on the 600-pair subset, the denominator of the dashed, dotted and dash-dotted prompted judges. The solid line is Opus 5.5 on all 2,892 pairs.}
\label{fig:size}
\end{figure}

Language models fine-tuned on the outcomes stay ahead of every prompted judge, even at one thirteenth of the size.
Table~\ref{tab:stage1} reports every baseline under one evaluator.
By our rule, the highest pair accuracy on all test pairs among (a)--(e), the strongest baseline is \StrongestBaseline{} at \StrongestMOne\%. The runner-up \RunnerUp{} is close behind, but our main comparison is against \StrongestBaseline{} alone.
Prompted commercial judges (f) and persona mixtures (g), run on the 600-pair subset, stay below the three LoRA baselines (c)--(e) there, while the sentence classifiers are not clearly ahead of them and (b) is below (a).
Four further exploratory runs ask whether more effort on the asked side closes the gap, with intervals in Table~\ref{tab:stage1} and Appendices~\ref{app:aux} and~\ref{app:u3}.
The Opus judge on all test pairs reaches \OpusFullMOne\%, slightly below (a), a margin the subset cannot resolve.
The same judge at high reasoning effort with a majority of five votes per pair draws level with (a) on the subset.
A grounded model 13 times smaller, Qwen3 0.6B base with the count-level objective, reaches \SizeSmallMOne\%, above every prompted judge (Fig.~\ref{fig:size}).
Exploratory checks keep its gain over Opus above zero for \ifnum\SizeSmallSeedN=3 all three\else \SizeSmallSeedN{}\fi{} seeds, \ifnum\SizeSmallSubLoN=\SizeSmallSubN all \SizeSmallSubN{}\else \SizeSmallSubLoN{} of \SizeSmallSubN{}\fi{} subsets and Llama 3.2 1B base (\OtherFamDiffOpus{} points, Table~\ref{tab:smallrobust}).
\UThreeBodyText

\subsection{Count-level against pair-level objective}
\label{sec:res-obj}
Training on the counts rather than on pairwise wins is our main test of grounding.
The count-level model reaches \AblMainMOne\% pair accuracy, \GateDiff{} points [\GateDiffCI] over the pair-level objective \StrongestBaseline{}.
\RegText{We have not scored the main comparison yet.}{The interval excludes 0, so the count-level objective meets our pre-registered criterion of beating the pair-level objective.}{The interval includes 0, so the count-level objective does not meet our pre-registered criterion of beating the pair-level objective. We can exclude gains above \GateCIHi{} points but not smaller ones.}
Of the two ablations, wins and losses instead of the binomial likelihood lower pair accuracy, and one shared $\alpha$ did not converge (Appendix~\ref{app:aux}).

\subsection{Feature diagnostic: prompted-model answers versus revealed clicks}
\label{sec:res-e2}
\begin{figure}[t]
\centering
\includegraphics[width=\linewidth]{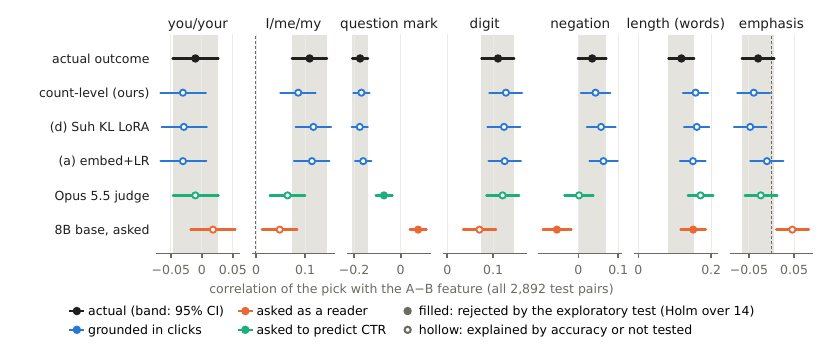}
\caption{Correlation of each predictor's pick with the A$-$B difference of seven headline features on all 2,892 test pairs, with 95\% bootstrap intervals. Black: the actual click outcome (gray band: its interval). Blue: trained on clicks (our count-level model and baselines (d) and (a)). Green: the Opus 5.5 judge (f), asked which got the higher CTR.\@ Orange: the 8B base model asked as a reader which headline they would click. A filled marker means that the exploratory test against a classifier blurred to the predictor's accuracy rejects it, Holm-corrected over the 14 tests of the asked predictors only.}
\label{fig:e2}
\end{figure}

A model asked which headline a reader would click does not reproduce the feature effects in the actual clicks, while the same model trained on the clicks does.
The clearest case is the question mark, a lean we call question-mark optimism.
In the actual outcomes a headline with a question mark loses more often than its sibling without one, with correlation \EtwoQActual{} in Figure~\ref{fig:e2}.
Qwen3-8B-Base asked as a reader picks the question headline more often, the opposite sign, and still does when asked which headline got the higher CTR, so its flip is the model's, not the wording's.
The Opus 5.5 judge keeps the sign but recovers less than half the effect, less than its accuracy predicts.
Sonnet 5.5, Haiku 5.5 and Qwen3-8B instruct, asked both questions on held-out test pairs whose answers we had not seen, also miss this penalty and are less accurate as readers.
The same base model with adapter (e), asked the same question, is at \EtwoQFt, inside the actual interval.
An exploratory test against a classifier blurred to each asked predictor's accuracy rejects the base model and Opus on the question mark (Appendices~\ref{app:e2cond} and~\ref{app:e2full}).
These show what each predictor weighs, not why readers clicked.

\begin{figure}[t]
\centering
\includegraphics[width=\linewidth]{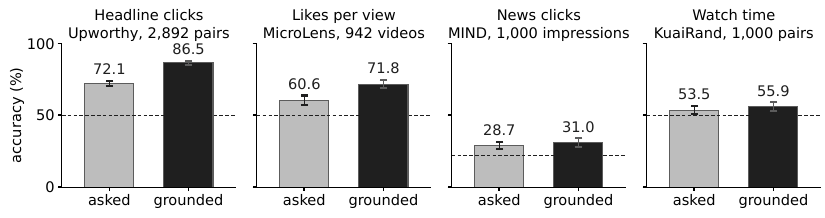}
\caption{Asked versus grounded across four behaviors without reader history: accuracy with 95\% intervals, or top-1 hit rate for news clicks. Dashed lines mark chance. Headlines: Opus 5.5 asked once versus the 8B grounded with (d), as in Figure~\ref{fig:overview}. Other panels: Sonnet asked versus the 8B grounded in that behavior, on the items asked (Table~\ref{tab:ext}).}
\label{fig:behaviors}
\end{figure}

\paragraph{Beyond headlines (Table~\ref{tab:ext}, Figure~\ref{fig:behaviors}, Appendix~\ref{app:behaviors}).} \ExtShortText{}
\input{tables/ext}

%% file: tables/u9numbers.tex
\newcommand{\UNineBodyText}{Scored zero-shot on another platform, adapters trained on one never beat asking on that platform under the registered metric (top-1 hit on MIND, pair accuracy elsewhere): in all six scorings the interval of transfer minus asked includes or lies below 0 (Table~\ref{tab:transfer}). The model trained on video likes falls below chance on Upworthy headlines (43.4\% [41.6, 45.3]).}
\newcommand{\UNineShortText}{Adapters do not carry over: scored zero-shot on another platform, none beats asking there in any of the six scorings (Table~\ref{tab:transfer}).}
\newcommand{\UNineAppText}{\paragraph{Zero-shot transfer across platforms (Table~\ref{tab:transfer}).} The comparison was registered before any adapter was scored on another platform and reuses adapters already trained; nothing was retrained. On MIND the asked 8B is below chance on top-1 (18.1\% and 19.7\% with the two prompt templates, against 22.1\%), so part of the Upworthy adapters' gain over the base model only undoes that deficit; against chance they add +5.4 and +4.3 points. On the secondary pair accuracy of MIND the pair-level adapter is +3.7 points above asking [+1.1, +6.2], but the registered verdict uses top-1, where its interval includes 0. The video-trained adapter is 5.9 points [3.6, 8.3] below the asked 8B on Upworthy pairs: what it learned from likes per view points the wrong way for headline clicks.}

%% file: tables/u3numbers.tex
\newcommand{\UThreeBodyText}{Given the training pairs as well, the asked model does not catch up: Sonnet 5.5 shown the ten most similar solved training pairs reaches 71.2\% on all 2,892 test pairs, far below (d).}
\newcommand{\UThreeAppText}{Two baselines dated after the objective comparison give the asked side the same 4,928 training pairs; they are descriptive and enter neither Table~\ref{tab:stage1} nor the strongest-baseline rule. \emph{Retrieval few-shot:} each test pair is preceded by the $k=10$ training pairs whose embedding differences $e(A)-e(B)$ (the vectors of (a)) have the largest absolute cosine to its own (mean largest 0.33), each with its actual winner, one pair per request to Sonnet 5.5 through the Claude Code CLI, which has no temperature option (default sampling). On all 2,892 test pairs it reaches 71.2\% [69.6, 72.9], +8.5 points [+6.4, +10.5] over the same model asked zero-shot (62.8\% [61.0, 64.5]), $-$3.5 points [$-$5.5, $-$1.4] from (a) and $-$15.3 points [$-$17.2, $-$13.5] from (d). The CLI invocation changed midway: 603 pairs (the 600-pair subset and 3 more) were answered with the CLI's default system prompt plus ours appended, as was every zero-shot answer, and the other 2,289 with our system prompt only; raw accuracy 69.5\% and 71.7\%, not tested. \emph{Calibrated regression:} a symmetric logistic regression without intercept on the base model's two asked logits (reader and CTR questions), with $C$ chosen by 5-fold cross-validation over 5 seeds on the training pairs, reaches 56.1\% [54.3, 57.9] on all test pairs, +2.5 points [+1.0, +3.9] over the reader question alone; adding Sonnet's zero-shot pick as a third feature gives 62.7\% [60.9, 64.5], $-$0.1 points [$-$0.3, +0.1] from that pick alone, $-$12.0 points [$-$14.2, $-$9.8] from (a) and $-$23.9 points [$-$25.8, $-$21.9] from (d).}

%% file: tables/stage1.tex
\begin{table*}[t]
\centering
\scriptsize
\setlength{\tabcolsep}{3pt}
\caption{Baselines and the count-level objective on the test period (2,892 pairs, 4,407 units); brackets: 95\% bootstrap intervals over experiments. M1 (pair accuracy) and M2 (top-1 hit rate within a unit) in \%, M3 in nats per 1,000 impressions (Section~\ref{sec:method}). $\Delta$: paired difference to (a) in points, rounded after differencing (so it can differ by 0.1 from the rounded cells); $^\dagger$: interval excludes 0. (f) and (g) were run on the 600-pair subset; the Opus 2,892-pair cells come from one auxiliary full run with the same prompt. The count-level row is the objective of Section~\ref{sec:method} (seed 0), not a baseline. Indented rows are auxiliary or reference settings, not candidates for the strongest baseline; (c$'$), the raw-CTR variant of (c), the Sonnet and Haiku judges and the other MoP settings are in the supplementary appendix to be released with the code{}.}
\label{tab:stage1}
\resizebox{\textwidth}{!}{%
\begin{tabular}{lllllll}
\toprule
Method & M1 (2,892) & M1 (600) & M2 & M3 & $\Delta$M1 (2,892) & $\Delta$M1 (600) \\
\midrule
(a) Embedding + logistic & 74.7 {[}73.1, 76.2{]} & 73.3 {[}69.7, 76.8{]} & 37.7 {[}36.3, 39.2{]} & 0.0585 {[}0.0523, 0.0649{]} & -- & -- \\
(b) TF-IDF + logistic & 69.4 {[}67.7, 71.1{]} & 68.8 {[}65.2, 72.3{]} & 33.7 {[}32.3, 35.1{]} & 0.0149 {[}0.0076, 0.0221{]} & $-$5.3 {[}$-$7.3, $-$3.3{]}$^\dagger$ & $-$4.5 {[}$-$9.0, $-$0.2{]}$^\dagger$ \\
(c) LOLA-style CTR regression & 82.7 {[}81.4, 84.1{]} & 81.8 {[}78.7, 84.8{]} & 43.4 {[}41.9, 44.9{]} & 0.0963 {[}0.0886, 0.1038{]} & +8.1 {[}+6.2, +9.9{]}$^\dagger$ & +8.5 {[}+4.3, +12.8{]}$^\dagger$ \\
(d) Suh-style option KL & 86.5 {[}85.3, 87.8{]} & 87.0 {[}84.3, 89.7{]} & 46.1 {[}44.6, 47.5{]} & 0.1216 {[}0.1123, 0.1308{]} & +11.9 {[}+10.3, +13.5{]}$^\dagger$ & +13.7 {[}+10.0, +17.3{]}$^\dagger$ \\
(e) Pairwise LoRA classifier & 85.5 {[}84.2, 86.8{]} & 85.7 {[}82.8, 88.5{]} & 45.4 {[}44.0, 46.9{]} & 0.0917 {[}0.0827, 0.1006{]} & +10.8 {[}+9.1, +12.5{]}$^\dagger$ & +12.3 {[}+8.5, +16.2{]}$^\dagger$ \\
Count-level objective (ours) & 87.4 {[}86.2, 88.6{]} & 86.8 {[}84.2, 89.5{]} & 47.0 {[}45.6, 48.5{]} & 0.1297 {[}0.1205, 0.1391{]} & +12.8 {[}+11.1, +14.5{]}$^\dagger$ & +13.5 {[}+9.8, +17.2{]}$^\dagger$ \\
\midrule
(f) Opus 5.5 judge & 72.1 {[}70.5, 73.8{]} & 70.2 {[}66.7, 73.8{]} & -- & -- & $-$2.6 {[}$-$4.7, $-$0.5{]}$^\dagger$ & $-$3.2 {[}$-$7.8, +1.7{]} \\
(g) MoP, Sonnet, fitted mixture & -- & 71.0 {[}67.5, 74.7{]} & -- & -- & -- & $-$2.3 {[}$-$6.8, +2.2{]} \\
\bottomrule
\end{tabular}}
\end{table*}

%% file: tables/ext.tex
\begin{table}[t]
\centering
\scriptsize
\setlength{\tabcolsep}{2pt}
\caption{Asked vs.\ grounded beyond headlines (\%, [95\% CI]). Asked = Sonnet via API (MIND: command line); MIND: top-1 hit rate, chance 21.8\%; MicroLens: past rate = video length; $\Delta$: vs.\ title only; new videos: registered test on earlier command-line answers. Rest: Table~\ref{tab:extfull}; Holm: Table~\ref{tab:holm}.}
\label{tab:ext}
\begin{tabular}{@{}l r cc cc cc@{}}
\toprule
 & & & & \multicolumn{2}{c}{Asked LLM} & \multicolumn{2}{c}{Grounded 8B} \\
\cmidrule(lr){5-6}\cmidrule(lr){7-8}
Data (behavior) & $n$ & Past rate & Embedding & plain & +history & plain & +history \\
\midrule
KuaiRand-Pure (watch time) & 3,000 & 53.6 {\tiny [51.8, 55.4]} & 55.2 {\tiny [53.4, 57.0]} & -- & -- & 56.2 {\tiny [54.5, 58.0]} & 56.8 {\tiny [55.0, 58.6]} \\
\quad subset & 1,000 & 53.4 {\tiny [50.3, 56.5]} & 56.1 {\tiny [53.0, 59.2]} & 53.5 {\tiny [50.7, 56.2]} & 59.4 {\tiny [56.7, 62.0]} & 55.9 {\tiny [52.7, 59.0]} & 56.1 {\tiny [53.0, 59.2]} \\
MIND-small (click) & 20,000 & 26.1 {\tiny [25.5, 26.7]} & -- & -- & -- & 28.9 {\tiny [28.3, 29.6]} & 38.2 {\tiny [37.5, 38.9]} \\
\quad subset & 1,000 & 26.1 {\tiny [23.6, 28.6]} & -- & 28.7 {\tiny [26.4, 31.2]} & 38.1 {\tiny [35.3, 40.8]} & 31.0 {\tiny [28.1, 33.9]} & 40.2 {\tiny [37.3, 43.2]} \\
\quad confirmation sample & 3,000 & -- & -- & -- & 38.6 {\tiny [36.8, 40.3]} & -- & 38.1 {\tiny [36.4, 39.9]} \\
MicroLens-100K (likes/view) & 942 & 68.6 {\tiny [65.6, 71.5]} & 66.9 {\tiny [63.7, 69.8]} & 60.6 {\tiny [57.4, 63.8]} & -- & 71.8 {\tiny [69.1, 74.5]} & -- \\
\quad + frames & 942 & -- & 73.9 {\tiny $\Delta$+7.0} & 61.7 {\tiny $\Delta$+1.1} & -- & 79.1 {\tiny $\Delta$+7.3} & -- \\
\quad new videos (title / + frames) & 200 & -- & -- & 59.0 / 58.0 & -- & 68.0 / 78.0 & -- \\
\bottomrule
\end{tabular}
\end{table}

%% file: appendix.tex
\ifdefined\ContamNItems\else\input{tables/contamnumbers}\fi
\ifdefined\AuxNPairs\else\input{tables/auxnumbers}\fi
\ifdefined\ReproModels\else\input{tables/repronumbers}\fi
\ifdefined\EtwoBaseAcc\else\input{tables/e2numbers}\fi
\ifdefined\AblK\else\input{tables/ablnumbers}\fi
\ifdefined\UNineAppText\else\input{tables/u9numbers}\fi
\ifdefined\BenchAppText\else\input{tables/benchnumbers}\fi
\ifdefined\UThreeAppText\else\input{tables/u3numbers}\fi

\FloatBarrier
\section{Contamination check}
\label{app:contam}
The archive is public, so the base model may have seen the headlines, and only winners were published on the site.
We gave the first half of the words of \ContamNItems{} held-out headlines (winner and loser of one pair in each of \ContamNTests{} test experiments) to the model and decoded greedily.
The base model completed the rest exactly for \ContamBaseExact\% of headlines [\ContamBaseExactCI] and the Suh-style adapter (d) for \ContamDExact\% [\ContamDExactCI] (Clopper--Pearson 95\% intervals, since these are counts of one and zero); the base model matched the first word for \ContamBaseFirst\% of headlines and, on average, \ContamBasePrefix\% of the remaining words before the first mismatch.
Winners were not recalled more often than losers (exact: \ContamBaseExactWL{} points [\ContamBaseExactWLCI]; leading-word share: \ContamBasePrefixWL{} [\ContamBasePrefixWLCI] for the base model and \ContamDPrefixWL{} [\ContamDPrefixWLCI] for (d)), so we see no sign that recall of published winners drives pair accuracy.
As an indirect bound on associative memory, the base model asked zero-shot which of two headlines a reader would click (the reader question of the feature diagnostic) picks the winner in \EtwoBaseAcc\% of the \ReproBaseAskN{} test pairs, so whatever it remembers about published winners gives it little without training.
The same check on the count-level model gives \ContamGroupExact\% exact completions [\ContamGroupExactCI], winner minus loser \ContamGroupPrefixWL{} points in leading-word share [\ContamGroupPrefixWLCI] and \ContamGroupFirstWL{} in first-word match [\ContamGroupFirstWLCI], and differs from the base model by \ContamGroupMinusBasePrefix{} [\ContamGroupMinusBasePrefixCI] and \ContamGroupMinusBaseFirst{} [\ContamGroupMinusBaseFirstCI] points on these two measures, so training on clicks did not add recall of winners.
The check tests verbatim recall only. It is post hoc: the registered rule did not include it.
We did not run the check on the commercial judges, whose training data may include the archive or the published winners. Such recall could only help them pick winners, so it cannot create the accuracy gap between asked and grounded models; it may shift their feature-diagnostic readings (Table~\ref{tab:e2}).

\FloatBarrier
\section{Auxiliary checks on the evaluation}
\label{app:aux}
None of these checks changes the registered comparison or the reported intervals; they help to read them.
\paragraph{Families of tests.} Table~\ref{tab:holm} lists each family of tests behind a verdict in the abstract or main text, whether it was registered, its size and what it rejects. Bootstrap $p$-values carry Monte Carlo error, $\sqrt{p(1-p)/B}$ for $B$ resamples. Moving every $p$ by \HolmMcKse{} such standard errors toward the opposite verdict leaves all decisions unchanged in \HolmMcFamStable{} of the \HolmMcFamWithP{} families with $p$-values (the objective comparison is decided by its interval). The exception is the post hoc feature-diagnostic family: \HolmMcETwoFlipN{} tests it does not reject (base 8B digits, Opus negation) lie within \HolmMcETwoFlipZ{} standard errors of their Holm thresholds, so its \EtwoGAskedN{} rejections should be read as \HolmMcETwoMin{} to \HolmMcETwoMax{} (post hoc, rule dated before the run).

\input{tables/holm}
\paragraph{Result ledger.} Table~\ref{tab:ledger} maps each claim to its status, verdict and section.

\input{tables/ledger}

\paragraph{Margins behind Sections~\ref{sec:res-stage1} and~\ref{sec:res-obj}.} The strongest baseline \StrongestBaseline{} is \RunnerUpDiff{} points [\RunnerUpDiffCI] ahead of the runner-up \RunnerUp{}, so its choice does not rest on a clear margin. The random top-1 hit rate is \MTwoRandom\%. Replacing the binomial likelihood with wins and losses changes pair accuracy by \AblWinlossDiff{} points [\AblWinlossDiffCI].
\paragraph{Margins behind Section~\ref{sec:discussion}.} \RegText{}{}{\AuxAllPairsText{} On the 8B, the two grounded objectives that fit click shares are within two points of each other and CTR regression (c) is about four points lower. A logistic probe on the frozen 8B (Frozen base-model features, below) reaches \AuxPrMOne\%, \AuxPrDiffA{} points of the \AuxPrDMOne--\AuxPrAMOne{} gap between (d) and (a).} \SeedText{}{Retrained with seeds 1 and 2, the count-level model averages \SeedGroupMOne\% [\SeedGroupMOneCI] and differs from \StrongestBaseline{} by \SeedDiff{} points [\SeedDiffCI] on the three-seed means (auxiliary; per-seed values in Table~\ref{tab:seeds}); resampling seeds as well as experiments widens this to [\SeedVarCI] (post hoc), which includes zero and is too wide to show equivalence within $\pm$1 point.}

\paragraph{Shared-$\alpha$ ablation.} The registered ablation with one shared $\alpha$ instead of $\alpha_i$ changes pair accuracy by \AblSharedDiff{} points [\AblSharedDiffCI], but that run did not converge (median final loss \AblSharedLoss{} vs.\ \AblMainLoss{} for the main run), so the number mixes losing $\alpha_i$ with failed optimization and is not read as the value of $\alpha_i$.

\paragraph{The 600-pair subset.} The subset used for the commercial judges is the pilot's draw, \AuxSampleRule{} over the \AuxNPairs{} test pairs, fixed before any method in this paper was run.
With the paired experiment-level bootstrap used throughout, a difference between two trained baselines on this subset is detectable at 80\% power and two-sided 5\% only if it exceeds \AuxMdeSixLo--\AuxMdeSixHi{} points (post hoc), so the subset separates only large gaps.
The reference settings of the commercial judges and of MoP that are not shown in Table~\ref{tab:stage1} are in the supplementary appendix to be released with the code{}.
The strongest prompting in Section~\ref{sec:res-stage1} (Opus 5.5, high reasoning effort, five votes per pair with the headline order drawn at random per vote, ties counted as wrong) reaches \AskStrongMOne\% (\AskStrongMOneCI) on the 600 pairs and is \AskStrongDiffA{} points from the (a) value of Table~\ref{tab:stage1} on the same 600 pairs [\AskStrongDiffACI] (post hoc, outside every Holm family).
The grounded 0.6B model of Section~\ref{sec:res-stage1} reaches \SizeSmallMOne\% (\SizeSmallMOneCI) on all 2,892 pairs and \SizeSmallMOneSix\% on the 600; it is \SizeSmallDiffOpus{} points above Opus 5.5's single judgment on all 2,892 pairs [\SizeSmallDiffOpusCI] and \SizeSmallDiffStrong{} points above this five-vote majority on the 600 [\SizeSmallDiffStrongCI] (same paired bootstrap, post hoc).
Retrained with \SizeSmallSeedN{} seeds in all, it reaches \SizeSmallSeedMin--\SizeSmallSeedMax\%, and every seed's interval for the gain over Opus 5.5's single judgment stays above zero (lowest lower end \SizeSmallSeedLoMin{} points; post hoc, rule registered before the extra seeds were trained).
On seed 0 the lower end of the interval stays above zero in \SizeSmallSubLoN{} of the \SizeSmallSubN{} test subsets of Table~\ref{tab:smallrobust} (creation month, terciles of the click-rate gap, the four formats; smallest gain \SizeSmallSubMinDiff{} points). Llama 3.2 1B base trained the same way reaches \OtherFamMOne\%, \OtherFamDiffOpus{} points above Opus 5.5's single judgment [\OtherFamDiffOpusCI] (post hoc, one seed).
\input{tables/smallrobust}
Figure~\ref{fig:size} adds Qwen3 1.7B and 4B base trained the same way (post hoc). At all four sizes both grounded objectives stay above every prompted judge, and the lower end of each interval stays above Opus 5.5's single judgment on all 2,892 pairs; from 4B to 8B the count-level model changes by \SizeCountFourEight{} points [\SizeCountFourEightCI], an interval that includes zero, while the pairwise classifier (e) gains \SizePairFourEight{} points [\SizePairFourEightCI] (paired bootstrap on the same pairs, post hoc, no multiplicity correction).
With Llama 3.1 8B base in place of Qwen3 8B base and the same count-level settings (post hoc, one seed), pair accuracy is \LlamaMOne\% (\LlamaMOneCI) on all 2,892 pairs, \LlamaDiffQwen{} points from the Qwen3 8B run [\LlamaDiffQwenCI] (paired bootstrap, an interval that includes zero).
\AuxDcText

\FloatBarrier
\section{Held-out formats and time shift}
\label{app:heldout}
Leaving a headline format out of training costs the retrained predictors little: the one format drop of the classifier (a) beyond noise is \EthreeNumDiff{} points, when digits are left out, and the loss from training on old data only is largely a loss of training data.
Whether the grounded LoRA model loses less than the classifier is undetermined at this power, since in no condition does its drop differ from that of (a) beyond noise.
Table~\ref{tab:e3} asks how much a predictor loses on headlines of a format it never saw in training.
For each of four formats we remove every training pair in which either headline has the format and evaluate on the test pairs that contain it; the time-shift condition trains only on pairs created before 2013-06-25 and evaluates on pairs from 2015 on, after Facebook's 2014 change against click-bait headlines~\citep{elarini2014clickbaiting}.
Leaving out question marks or first-person headlines costs (a) nothing measurable; the larger quotation drop rests on 325 pairs and its interval includes zero.
With one seed per condition only drops of a few points are detectable.
The time shift costs (a) \EthreeTimeDiff{} points, its one drop that survives Holm, but an equally sized random draw of recent pairs (post hoc) already costs \EthreeSizePart{} points, and the further loss due to the age of the data is within noise.
Adapter (c) loses beyond noise only under the time shift.
\input{tables/e3}

\begin{figure}[ht]
\centering
\includegraphics[width=\linewidth]{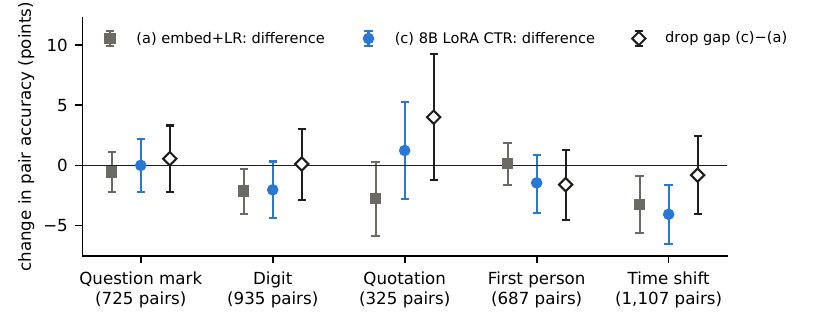}
\caption{Held-out-format drops (Table~\ref{tab:e3}): change in M1 on the target pairs when training leaves the format out (time shift: trains on old data only), held-out minus full training, in points, with 95\% paired bootstrap intervals. Gray squares: classifier (a); blue circles: adapter (c) retrained on the same reduced data; open diamonds: drop gap, (c) difference minus (a) difference (positive: (c) loses less). Seed 0, as registered.}
\label{fig:e3drops}
\end{figure}

Figure~\ref{fig:e3drops} plots the differences of Table~\ref{tab:e3} with their intervals.

\paragraph{Power of the LoRA side of the held-out-format comparison (post hoc).} \AuxLoraMdeLead{}, a difference between the LoRA drop and the classifier drop is detectable at 80\% power only if it exceeds about \AuxLoraMdeQuestion{} (questions), \AuxLoraMdeNumber{} (numbers), \AuxLoraMdeQuote{} (quotes), \AuxLoraMdeFirst{} (first person) and \AuxLoraMdeTime{} points (time shift). Per condition the drop-gap intervals exclude (c) losing less than (a) by more than \EthreeGapLessMin--\EthreeGapLessMax{} points, and (c) losing more than (a) by more than \EthreeGapMoreMin--\EthreeGapMoreMax{} points.

\FloatBarrier
\section{Feature diagnostic: full table}
\label{app:e2cond}
Table~\ref{tab:e2} gives the feature diagnostic of Section~\ref{sec:res-e2} for every feature and predictor on all test pairs. Errors that ignore the features shrink correlations toward 0 but cannot flip a sign. The main post hoc test of Section~\ref{sec:res-e2} therefore compares each asked predictor with classifier (a) blurred by such errors to that predictor's accuracy ($G$, Holm-adjusted over the \EtwoGAskedM{} tests of the base model and Opus). It rejects the base model on question mark, negation and length and Opus on question mark (\EtwoGAskedN{} of \EtwoGAskedM; \HolmMcETwoMin{} to \HolmMcETwoMax{} within Monte Carlo error, Table~\ref{tab:holm}). The base model flips the sign on question mark, negation and emphasis, though only the question mark has a nonzero effect in the clicks; the fine-tuned model asked, which is more accurate than (a), and every click-trained predictor flip none.

\input{tables/e2}

The parentheses of Table~\ref{tab:e2} give a sensitivity reading, $D$: the excess over a predictor of equal pair accuracy whose errors ignore the features, which needs no blurred classifier. $G$ is the reading with the fewest rejections, and the base model's question mark and negation are rejected under all three readings ($G$, $D$ and a conditional regression on the features and the CTR gap). The other readings, the conditional checks and a replication of the actual side on all pairs of headlines within the test experiments are in a supplementary appendix to be released with the code.

\paragraph{Margins behind Section~\ref{sec:res-e2}.} On question mark the base model asked as a reader has pick correlation \EtwoQBase{} and the Opus judge \EtwoQOpus{} (Table~\ref{tab:e2}); the Opus value is below what its accuracy predicts, $G=\EtwoGOpusQ$ [\EtwoGOpusQCI]. Pair accuracies on the full test set are \EtwoBaseAcc\% for the base model asked, \EtwoOpusAcc\% for Opus and \EtwoFtAcc\% for the fine-tuned model asked; the last is above (a), so it is outside the $G$ family of \EtwoGAskedM{} tests (base model and Opus, seven features), in which \EtwoGAskedN{} reject. A sign flip means that the pick correlation's interval excludes 0 with the sign opposite to actual. The base model's CTR-question run and the three models of Appendix~\ref{app:e2full} give the other numbers of that section (Appendix~\ref{app:prompts}: \EtwoBaseTwoQJudgeQ{} on question mark; \EtwoGTenQN{} of six arms rejected by $G$ on question mark).

\FloatBarrier
\section{Asking three models two questions}
\label{app:e2full}
Table~\ref{tab:e2full} asks Sonnet 5.5, Haiku 5.5 and Qwen3-8B (instruct) both questions of Appendix~\ref{app:prompts} on every registered test pair, that is, the \EtwoFullN{} test pairs outside the \EtwoFullNSix-pair subset whose answers had been seen, in both orders. The rule was dated after the objective comparison was scored, so these rows are an addition and do not enter Table~\ref{tab:stage1}. In the clicks a question mark loses (\EtwoFullQActual); every model and question stays between \EtwoFullQMin{} and \EtwoFullQMax, and all six raw asked$-$actual differences on this feature are rejected (\EtwoFullHolmAN{} of \EtwoFullHolmAM{} in the family of these six arms). Against the accuracy-matched null $G$ of Section~\ref{sec:res-e2}, \EtwoGTenQN{} of the six are rejected (Sonnet with both questions, Haiku and Qwen asked about CTR) and \EtwoGTenN{} of \EtwoGTenM{} overall. Asked as readers, all three are less accurate than asked about CTR (\EtwoFullDiffMin{} to \EtwoFullDiffMax{} points; Holm over the reader$-$CTR feature differences of the three models rejects \EtwoFullHolmBN{} of \EtwoFullHolmBM). Qwen3-8B ran 4-bit quantized, not the bf16 original. Asked as a reader it gives the same pick in both orders on only \EtwoFullQwenReaderAgree\% of pairs, so most of its picks are 0.5 and its correlations carry little information; this arm is part of \EtwoFullHolmAQwenReader{} of the \EtwoFullHolmAN{} rejections in the first family and \EtwoFullHolmBQwen{} of the \EtwoFullHolmBN{} in the second, which should be read with that caution. Scored by the logit difference instead of the pick it reaches \EtwoFullQwenLogitCtr\% (CTR) and \EtwoFullQwenLogitReader\% (reader).

\input{tables/e2full}

\FloatBarrier
\section{Low click rates: the profiled likelihood is multinomial}
\label{app:multinomial}
At low click rates $\sigma(x)\approx e^{x}$ and $\log(1-r)\approx -r$, so with $r_{ij}=\exp(\alpha_i+s_{ij})$ the unit's binomial log-likelihood~\eqref{eq:ll} is approximately the Poisson one, $\mathcal{L}_i\approx\sum_j c_{ij}(\alpha_i+s_{ij})-n_{ij}e^{\alpha_i+s_{ij}}$.
Its maximum over $\alpha_i$ is at $e^{\hat\alpha_i}=C_i/\sum_j n_{ij}e^{s_{ij}}$ with $C_i=\sum_j c_{ij}$, and substituting gives
\begin{equation*}
\begin{aligned}
\mathcal{L}_i(\hat\alpha_i)&\approx\sum_j c_{ij}\log\operatorname{softmax}_j\!\big(s_{ij}+\log n_{ij}\big)+\mathrm{const}\\
&=-C_i\,\mathrm{KL}\!\big(q_i\,\big\|\,\operatorname{softmax}(s_i+\log n_i)\big)+\mathrm{const},
\end{aligned}
\end{equation*}
where $q_{ij}=c_{ij}/C_i$ and the constants do not depend on the model.
Baseline (d) minimizes $\mathrm{KL}(q^{d}_i\,\|\,\operatorname{softmax}(t_i))$ with $q^{d}_{ij}\propto c_{ij}/n_{ij}$ and option-letter logits $t_{ij}$.
When the $n_{ij}$ of a unit are equal, $q^{d}_i=q_i$ and $\log n_{ij}$ cancels in the softmax, so the two unit objectives agree up to the factor $C_i$; with unequal impressions both are still minimized by free scores equal to $\log(c_{ij}/n_{ij})$ up to a constant.
What remains different is the weight of a unit ($C_i$ against 1), the exact $\sigma$ and binomial terms, and how the scores are produced: our $s_{ij}$ comes from a prompt with one headline, (d)'s $t_{ij}$ from one prompt listing up to eight headlines of the unit.
For the count-level objective, where $s_{ij}=\ell(h_{ij})$, the comparison with (d) is therefore close to a comparison of a pointwise and a listwise parameterization of the same multinomial likelihood.

\FloatBarrier
\section{Broader impacts}
\label{app:impacts}
Predicting real clicks lets editors pre-test wording without exposing readers to every variant, and the asked-versus-revealed diagnostic warns against trusting simulated readers. The same model could optimize click-bait, the target of Facebook's 2014 ranking change~\citep{elarini2014clickbaiting}; we predict clicks, not reader benefit or quality. The data are aggregate counts, so the models cannot profile or target a person; the two Upworthy adapters to be released{} (Appendix~\ref{app:repro}) carry model cards that state this intended use.

\FloatBarrier
\section{Behaviors covered}
\label{app:behaviors}
Table~\ref{tab:behaviors} lists the behaviors this paper considers: clicks, short-video watch time and short-video likes per view. Only the Upworthy row comes from randomized A/B tests. The KuaiRand~\citep{gao2022kuairand} test pairs come from videos inserted at random into a production feed (its training log is recommended), and the MIND log~\citep{wu2020mind} and the MicroLens totals~\citep{ni2023microlens} are shaped by a recommender, so their labels mix item appeal with exposure. A further short-video set, SnapUGC, was considered and dropped because its terms restrict use to a 2025 prediction contest.

\input{tables/behaviors}
\paragraph{Margins behind ``Beyond headlines''.} \ExtBodyText{} \AudienceText{} \MindAbsText{}
\paragraph{Watch time (KuaiRand).} On short-video watch time, \KwAccText. For the audience-conditioned variants, \KwAskedCondText{}. \KwMdeText

\input{tables/ext_full}
\paragraph{News clicks (MIND).} \MindAppText{} \MindMdeText
\paragraph{Frame descriptions (MicroLens).} \MlFramesAppText
\paragraph{Power of the MicroLens checks (post hoc).} \MlMdeText
\paragraph{Title length (MicroLens, post hoc).} \MlLengthText

\FloatBarrier
\section{Limits of the logged data and of having no human survey}
\label{app:obslimits}
\paragraph{Logged exposure.} Within one Upworthy test the article and image were fixed and headlines were assigned at random, so a pair difference reflects wording alone. The other rows of Table~\ref{tab:behaviors} lack this property. In a recommender log, whether an item was shown, where in the list and to whom depend on the recommender's earlier estimate of its appeal, so an item's click or watch rate mixes its appeal with its exposure, its creator's audience and the period in which it circulated. KuaiRand narrows this for evaluation only: its test pairs come from random insertions and pair two videos of the same user, day and length bin, but the past rates and viewing histories used as inputs come from the recommended standard feed. MIND impression lists are recommended throughout and MicroLens like and view totals accumulate under the platform's recommender, so a model fitted to them may learn what the recommender chose to show. We therefore treat these datasets as checks of whether the asked-versus-grounded pattern recurs beyond headlines, not as tests of exposure effects.
\paragraph{No human survey.} No person was asked anything in this study. ``Asked'' always refers to prompted models, and ``revealed'' only to the logged clicks and watch time used for training and evaluation. Whether people's own answers to the pairwise question depart from their clicks in the same direction as the models' answers is therefore not tested here. A survey that puts the same pairs to a sample of readers would answer it, but it needs ethics review and cannot reach the 2013--2015 social-feed audience behind the archive; we leave it to future work.

\FloatBarrier
\section{Training and baseline details}
\label{app:training}
The base model is Qwen3-8B-Base, run with MLX on one Apple M5 Max (64\,GB).\@
LoRA adapters of rank 16 (scale 2.0, no dropout) are attached to the linear layers of all transformer blocks; the base weights are frozen.
\SeedText{All LoRA runs use AdamW without weight decay, 20 warm-up steps from 1\% of the peak learning rate followed by cosine decay to 10\% of it, gradient clipping at 1.0, two epochs and seed 0.}{All LoRA runs use AdamW without weight decay, 20 warm-up steps from 1\% of the peak learning rate, cosine decay to 10\% of it, clipping at 1.0, two epochs and seed 0 (also 1, 2; Section~\ref{sec:discussion}).}
The peak learning rate was chosen among $10^{-5}$, $3\times10^{-5}$ and $10^{-4}$ on the development set for (c), (d) and (e) separately ($10^{-5}$ was added later; the supplementary appendix to be released with the code{}); all three picked $3\times10^{-5}$, which we then fixed for our method.
The final (d) run used batch 8 instead of the batch 16 of its selection runs (the supplementary appendix to be released with the code{}), so we repeated its selection at batch 8 with the same three rates and the same number of examples (registered before running, development set only); the batch-8 reselection picks the same learning rate ($3\times10^{-5}$) as the batch-16 selection used for the final run.
For the count-level model the learnable prompt vectors use a 30$\times$ larger learning rate (same schedule shape).
The multiplier was chosen among 1, 30 and 300 in three development runs of an extension not reported here (protocol file; values in the supplementary appendix to be released with the code{}).
Each optimization step of the count-level model processes a batch of 16 units.
\SeedText{Click logits are computed in float32 since bfloat16 logits often tie.}{Click logits use float32 since bfloat16 logits often tie.}
\SeedText{}{Table~\ref{tab:seeds} gives the per-seed test accuracy of the four LoRA models behind the objective comparison.}

\input{tables/seeds}

\paragraph{Mixture-of-Personas baseline (g).} Eight personas are written by Sonnet 5.5 from training-pair clusters (KMeans on headline embeddings) with their winners marked; each persona answers with eight in-context examples from its cluster. The persona answers are combined by a gate $\operatorname{softmax}_k(\phi_k+\tau\cos(e(x),\mu_k))$ with per-persona error rates, all fitted by likelihood on 800 training pairs. Run with Haiku and Sonnet 5.5 on the 600-pair subset.

\paragraph{Prompt baselines given the training pairs (post hoc).}\label{app:u3} \UThreeAppText

\FloatBarrier
\section{Prompts}
\label{app:prompts}
Every prompt below is the exact string the code sends; parts in braces are filled per request. The prompts and training runs of the three extension datasets are in Appendix~\ref{app:exttrain}.
The word ``reader'' in the training prompt ``Would this reader click'' and in the initialization string ``Reader profile:'' (Appendix~\ref{app:repro}) is literal text from the code; it names no group of readers, and no model in this paper has reader-specific parameters.
In (g), \{persona description\} is the text that the persona-writing prompt returned for that persona, and \{exemplar pairs\} is that persona's in-context training pairs, one per line in the labeled format shown.
The commercial judges (f) receive ten pairs per request in the user message, one numbered pair per line in the format shown, with X and Y assigned in a pair-specific random order; the persona mixtures (g) use the same user message after the system message of their persona.
For the reader question of the feature diagnostic the model's next-token logits for `` A'' and `` B'' after ``Answer:'' are read off, and each pair is asked in both orders; the score is half the difference between the A$-$B logit gap in the original order and in the swapped order, which removes a constant preference for the first or second position.
Asked the CTR question instead (post hoc: same frame, first sentence and question taken from the judge prompt of (f)), the same base model picks the winner in \EtwoBaseTwoQJudgeAcc\% of the test pairs against \EtwoBaseTwoQReaderAcc\% as a reader (difference \EtwoBaseTwoQAccDiff{} [\EtwoBaseTwoQAccCI]), but its question-mark correlation stays positive (\EtwoBaseTwoQJudgeQ{}; reader \EtwoBaseTwoQReaderQ{}, actual \EtwoBaseTwoQActualQ). Negation is different: as a reader the base model avoids negated headlines (\EtwoBaseTwoQReaderNeg), but asked the CTR question it does not (\EtwoBaseTwoQJudgeNeg; actual \EtwoBaseTwoQActualNeg; reader minus CTR question \EtwoBaseTwoQNegDiff{} [\EtwoBaseTwoQNegCI], Holm over seven axes), so its negation gap in the main text is partly the wording's.

The system message of the commercial judges (f) names Upworthy and 2014--2015, so a judge could answer from remembered test outcomes rather than from the headlines; we did not run a variant without these names, so this retrieval path is unchecked (the completion check of Appendix~\ref{app:contam} covers the 8B model only).

\input{tables/prompts}

\input{tables/extrain}

\FloatBarrier
\section{Pilot timeline and reproduction details}
\label{app:repro}
The pilot saw test accuracies before the protocol was written; a supplementary appendix to be released with the code{} lists, with commit times, what it saw and what was fixed afterwards.

\input{tables/repro}
\paragraph{Run log and initialization.} This paragraph adds to Appendix~\ref{app:training}. Table~\ref{tab:repro} gives the prompt string and optimizer settings of each final run, and Table~\ref{tab:reprorun} lists what each final run fitted and how long it took (\ReproTrainMinTotal{} hours of training for the four runs). AdamW uses $\beta=(\ReproBetaOne,\ReproBetaTwo)$ and $\epsilon=\ReproEps$. Inputs are not truncated: each batch is right-padded to its longest prompt and the last real token is read. The (c) head is a linear layer initialized at zero, applied to the last hidden state scaled to unit RMS.\@ In the count-level model the learnable prompt has \ReproPrefixLen{} vectors that start at the embeddings of ``\texttt{ Reader profile:}'' plus Gaussian noise with standard deviation 0.1 times their RMS, and each $\alpha_i$ is profiled out by Newton steps rather than stored.
\paragraph{Protocol amendments.} The protocol file has \ReproPrN{} dated commits: \ReproPrPzero{} before any baseline result, \ReproPrPone{} after baseline results were seen but before any count-level test score, none between the first count-level test score and the objective comparison, and \ReproPrPthree{} after that comparison. By kind, \ReproPrRule{} fix registered rules (\ReproPrPoneRule{} of them after baseline results were seen), \ReproPrClaim{} lower or reword claims (one drops an interpretability claim of the extension, and three are the scope cuts below), \ReproPrAux{} register auxiliary analyses before they were run, \ReproPrPost{} add post hoc analyses, and \ReproPrRecord{} correct records or timestamps without changing a rule. None of the \ReproPrPthree{} commits after the objective comparison changes a rule: \ReproPrPthreePost{} add post hoc analyses, \ReproPrPthreeAux{} register auxiliary analyses before they were run, \ReproPrPthreeRecord{} correct records, and \ReproPrPthreeClaim{} narrow the reported scope: after the comparison the authors removed from this paper, in three commits, an extension that mixes several count-level adapters, a KuaiRand check that predicts random-exposure view rates from logged feeds, and a four-dataset summary score with a zero-shot transfer check across platforms; their registered rules are unchanged, and their code and results will be released together with a supplementary appendix that holds the paragraphs and tables moved out of this edition. Of all entries, \ReproPrWrittenAfter{} were written after the run they describe and say so. The power analysis of the objective comparison (Appendix~\ref{app:aux}) was added on \ReproPrMdeDate{}, after the baselines had been scored, so its minimum detectable difference of \AuxMdeProxy{} points is post hoc. The rule that the objective comparison is judged on seed 0 and that seeds 1 and 2 are auxiliary was committed on 13 Nov 2026 at 08:14 KST, a day before the first seed-1 run started (14 Nov, 16:07). Our count-level model also had more development runs than the baselines (the supplementary appendix to be released with the code{}), which favors it and so cannot explain a null result. We specified the baseline comparisons partly after seeing baseline results, so they are less protected than the objective comparison. %
\input{tables/amend}Table~\ref{tab:amend} lists the \AmendN{} rule changes made after baseline results were seen. \AmendCand{} set how a candidate for the strongest baseline is scored; each of those candidates ended at least \AmendCandGap{} points of pair accuracy below (d) ((f) and (g) on the 600-pair subset), so none of these changes decided which baseline is strongest. \AmendModel{} fix how the count-level model is trained, chosen or entered into the comparison, all before any count-level test score. The other \AmendOther{} concern ablations, the unreported extension, the KuaiRand analysis, seeds and run order.
\par{\rightskip=0pt plus 4em\relax \paragraph{Software and model versions.} The base model is the Hugging Face revision 49e3418 of Qwen3-8B-Base; \SuppRef{} \CodeRepoParen{} lists every Python package version (\texttt{results/env\_freeze.txt}, MLX 0.32.4, mlx-lm 0.32.2) and Table~\ref{tab:reprorun} gives the rows per stage. The code will be{} released under the MIT License.
Every table and in-text number is regenerated from the committed results by \texttt{bash scripts/reproduce\_cpu.sh} (CPU only, no model or API calls, about ten minutes on the machine of Table~\ref{tab:reprorun}), which ends by comparing each output byte for byte with the committed file. The prompt texts of Appendices~\ref{app:prompts} and~\ref{app:exttrain} are read with \texttt{ast} from the functions and constants that send them in the released code: \texttt{llm\_runs.py} (judges and pair line), \texttt{mop\_runs.py} (persona mixtures), \texttt{e2\_ask.py} (asked 8B), \texttt{lora\_common.py} and \texttt{kuairand\_watch\_asked.py} (KuaiRand), \texttt{mind\_lora\_text.py} and \texttt{mind\_level2\_asked.py} (MIND), \texttt{microlens\_lora\_text.py} and \texttt{microlens\_asked.py} (MicroLens). In MLX training the seed fixes the LoRA initialization (\texttt{mx.random.seed}), the order of training units and the noise of the initial prompt vectors (NumPy generators); we did not check that a rerun with the same seed gives identical weights.
\paragraph{Released adapters.} The count-level and pair-level (d) adapters of the registered comparison will be released under CC BY 4.0, each with a model card, its training command and package versions, and its predictions for the 2,892 test pairs. Inside a downloaded folder, \texttt{python score.py pairs} recomputes the card's pair accuracy and interval in seconds without the base model; \texttt{python score.py units -{}-base <Qwen3-8B-Base>} rescoring all test units with mlx-lm takes under half an hour on the machine of Table~\ref{tab:reprorun}. Adapters trained on MIND and MicroLens are not released, because their licenses reserve redistribution rights (MIND: Microsoft Research License Terms, Sec.~2--3; MicroLens: academic peer use, no distribution without explicit permission). %
Per-pair test scores of every baseline and grounded model (\texttt{results/*pair\_scores*.npy}) are in \SuppRef{} with the test pairs, so the pair accuracies and intervals can be recomputed without running any model.\par}
\paragraph{Pair selection.} Inside each experiment we keep the pairs of headlines that share image and lead paragraph and whose click rates differ under a two-sided two-proportion $z$-test at $p<0.05$, and then keep the one pair with the smallest $p$ (the ``most clearly separated'' pair); the order of the two headlines is flipped at random with seed 0.
\par{\rightskip=0pt plus 4em\relax \paragraph{API models.} The commercial judges and (g) were called through OpenRouter with the model identifiers \ReproModels. Their results were committed at \ReproMopDate{} ((g), 600 pairs), \ReproJudgeDate{} ((f), 600 pairs) and \ReproOpusFullDate{} (Opus 5.5 on all test pairs), all KST; the pilot calls were made on 11 Nov 2026. From 16 Nov 2026 on, Claude responses for the extension experiments are obtained through the Claude Code command-line interface on a subscription instead, with the same fixed prompt text and all tools disabled; the CLI (Claude Code \ReproCliVersion{} for every call) prepends its own system prompt and uses its default sampling settings, so these responses come through a different access path than the OpenRouter results above, and the two paths are never mixed within one comparison of answers (the MicroLens frame descriptions are model inputs, not compared answers, and use both; Appendix~\ref{app:behaviors}). On the same \ReproCliPairs{} judge pairs, Haiku 5.5 through the CLI scores \ReproCliRegAcc\% against \ReproCliRegOrAcc\% through OpenRouter under the registered metric (unanswered pairs count as wrong; \ReproCliNoAns{} unanswered pairs through the CLI, \ReproCliOrNoAns{} through OpenRouter), a paired difference of \ReproCliRegDiff{} points [\ReproCliRegDiffCI]; under the pilot rule that counts unanswered pairs as picking headline B the scores are \ReproCliAcc\% and \ReproCliOrAcc\% (\ReproCliDiff{} [\ReproCliDiffCI]); the CLI declined to answer \ReproCliRefused{} of the \ReproCliBatches{} batches and prefaced \ReproCliPreamble{} with a disclaimer, so asked results from the two paths are not compared across datasets. \ApiPathAppText{}\par}

\FloatBarrier
\section{Further related work}
\label{app:morerelated}
\paragraph{Generating for clicks.}
Industrial systems optimize generated creatives against click models~\citep{yang2024cg4ctr,lu2026onesize}.
A behavior-fitted simulator could serve as that reward model (untested).

\paragraph{Training and questioning simulators.}
OdysSim~\citep{zhou2026odyssim} trains an open 8B behavioral foundation model on 21.4M interactions and scores it on conversational and social tasks, not on logged click counts.
Asking a model and watching it act also come apart within LLMs themselves: their self-reports predict their own behavior only selectively, and persona prompts make self-reports more consistent without aligning behavior~\citep{kocielnik2026selfreports}; Section~\ref{sec:res-e2} compares asked and grounded readers against human clicks instead.

\FloatBarrier

%% file: tables/holm.tex
\begin{table}[ht]
\centering
\scriptsize
\setlength{\tabcolsep}{3pt}
\caption{Every family of tests behind a verdict in the abstract or main text, with its size $m$ and how many tests Holm's step-down rejects at $\alpha=0.05$ (a family of one has no adjustment). Registered: family and rule fixed before the run; post hoc: added after results were seen, reported as auxiliary. The confirmation families differ in size because MIND registered one condition ($m = 1$) and MicroLens two, title only and with frames ($m = 2$). The MicroLens rows use the registered command-line answers; Appendix~\ref{app:repro} (API models) gives the API re-runs. Other appendix tables state their own families in their captions.}
\label{tab:holm}
\begin{tabular}{@{}p{3.2cm}lp{5.0cm}rp{2.6cm}@{}}
\toprule
Family & Status & Tests in the family & $m$ & Rejected \\
\midrule
{\raggedright Objective comparison (Upworthy, 2,892 pairs)\par} & registered & {\raggedright count-level $-$ pair-level (d), M1\par} & 1 & {\raggedright none\par} \\
{\raggedright Held-out formats, classifier (a)\par} & registered & {\raggedright drop when one format is left out (4) or time shift\par} & 5 & {\raggedright 1: time shift\par} \\
{\raggedright Held-out formats, adapter (c)\par} & registered & {\raggedright drop when one format is left out (4) or time shift\par} & 5 & {\raggedright 1: time shift\par} \\
{\raggedright Feature diagnostic, asked predictors\par} & post hoc & {\raggedright feature correlation vs.\ blurred classifier ($G$), base 8B and Opus $\times$ 7 features\par} & 14 & {\raggedright 4: base 8B question mark, base 8B negation, base 8B length, Opus question mark\par} \\
{\raggedright Watch time (KuaiRand)\par} & registered & {\raggedright 8B $-$ asked Sonnet; 8B $-$ embedding; 8B $-$ popularity; 8B history $-$ 8B\par} & 4 & {\raggedright none\par} \\
{\raggedright News clicks (MIND)\par} & registered & {\raggedright 8B $-$ asked Sonnet (history, 1,000); 8B history $-$ 8B; 8B history $-$ popularity\par} & 3 & {\raggedright 2: 8B history $-$ 8B, 8B history $-$ popularity\par} \\
{\raggedright MIND confirmation (3,000 new)\par} & registered & {\raggedright 8B history $-$ asked Sonnet (history), top-1\par} & 1 & {\raggedright none\par} \\
{\raggedright MicroLens, frames (942 videos)\par} & registered & {\raggedright with frames $-$ title only, for asked Sonnet, embedding, 8B\par} & 3 & {\raggedright 2: embedding, 8B\par} \\
{\raggedright MicroLens confirmation (200 new)\par} & registered & {\raggedright 8B $-$ asked Sonnet, title only (T) and with frames (F)\par} & 2 & {\raggedright 2: title only, with frames\par} \\
{\raggedright LoRA vs.\ prompting (600 pairs)\par} & post hoc & {\raggedright 3 adapters (c, d, e) $-$ 7 prompt methods (f, g), M1\par} & 21 & {\raggedright all 21\par} \\
\bottomrule
\end{tabular}
\end{table}

%% file: tables/ledger.tex
\begin{table}[ht]
\centering
\scriptsize
\setlength{\tabcolsep}{3pt}
\caption{Result ledger: each claim, whether its test was registered, the verdict and where it is reported. Counts are Holm rejections in the families of Table~\ref{tab:holm}.}
\label{tab:ledger}
\begin{tabular}{@{}p{4.6cm}p{2.4cm}p{4.2cm}l@{}}
\toprule
Claim & Status & Verdict & Where \\
\midrule
{\raggedright Count-level objective beats the strongest baseline\par} & {\raggedright registered\par} & {\raggedright not met (interval includes 0)\par} & Sec.~\ref{sec:res-obj} \\
{\raggedright Asked answers depart from clicks by feature (claim 2)\par} & {\raggedright table registered; test post hoc\par} & {\raggedright 4 of 14 feature tests rejected\par} & Sec.~\ref{sec:res-e2} \\
{\raggedright Drop when a format or period is left out (held-out formats)\par} & {\raggedright registered\par} & {\raggedright classifier 1 of 5, adapter 1 of 5 rejected\par} & App.~\ref{app:heldout} \\
{\raggedright Grounded beats asked: watch time (KuaiRand)\par} & {\raggedright registered\par} & {\raggedright not met\par} & Tab.~\ref{tab:extfull} \\
{\raggedright Grounded beats asked: news clicks (MIND)\par} & {\raggedright registered\par} & {\raggedright not met; confirmation not confirmed\par} & Tab.~\ref{tab:extfull} \\
{\raggedright Frame text helps (MicroLens)\par} & {\raggedright registered\par} & {\raggedright 2 of 3 methods rise\par} & Tab.~\ref{tab:extfull} \\
{\raggedright Grounded beats asked: short videos (MicroLens)\par} & {\raggedright registered\par} & {\raggedright confirmed on new videos with command-line answers; not with API answers (title) or lengths matched\par} & Tab.~\ref{tab:extfull} \\
{\raggedright LoRA models beat prompting (600 pairs)\par} & {\raggedright post hoc\par} & {\raggedright 21 of 21 rejected\par} & Sec.~\ref{sec:res-stage1} \\
\bottomrule
\end{tabular}
\end{table}

%% file: tables/smallrobust.tex
\begin{table}[ht]
\centering
\scriptsize
\setlength{\tabcolsep}{3pt}
\caption{Robustness of the small grounded model (exploratory, rules fixed before the extra runs). Pair accuracy M1 (\%) of the count-level model with the 0.6B settings and of Opus 5.5's single judgment on the same pairs, and the gain with a paired 95\% bootstrap interval over experiments. Seeds and Llama 3.2 1B base use all 2,892 test pairs. Subsets use seed 0: creation month, terciles of the click-rate gap between the two headlines, and pairs where either headline has the format. Unanswered pairs count as wrong.}
\label{tab:smallrobust}
\begin{tabular}{lllll}
\toprule
Method & Pairs & Grounded & Opus 5.5 & Gain [95\% CI] \\
\midrule
Qwen3 0.6B, seed 0 & 2{,}892 & 83.2 & 72.1 & +11.1 {[}+9.2, +13.1{]} \\
Qwen3 0.6B, seed 1 & 2{,}892 & 82.7 & 72.1 & +10.7 {[}+8.7, +12.6{]} \\
Qwen3 0.6B, seed 2 & 2{,}892 & 82.5 & 72.1 & +10.4 {[}+8.5, +12.3{]} \\
Seed 0, Created: 2014-10 & 771 & 82.4 & 72.2 & +10.1 {[}+6.5, +13.9{]} \\
Seed 0, Created: 2014-11 & 556 & 82.6 & 71.2 & +11.3 {[}+6.8, +16.0{]} \\
Seed 0, Created: 2014-12 & 458 & 83.4 & 71.2 & +12.2 {[}+7.2, +17.2{]} \\
Seed 0, Created: 2015-01 & 359 & 83.8 & 74.4 & +9.5 {[}+4.2, +15.0{]} \\
Seed 0, Created: 2015-02 & 281 & 87.2 & 74.0 & +13.2 {[}+6.8, +19.6{]} \\
Seed 0, Created: 2015-03 & 241 & 84.2 & 69.7 & +14.5 {[}+7.5, +21.6{]} \\
Seed 0, Created: 2015-04 & 226 & 80.5 & 72.1 & +8.4 {[}+2.2, +14.6{]} \\
Seed 0, Click-rate gap: low & 964 & 79.8 & 67.3 & +12.4 {[}+8.9, +16.0{]} \\
Seed 0, Click-rate gap: mid & 964 & 82.8 & 71.5 & +11.3 {[}+7.9, +14.7{]} \\
Seed 0, Click-rate gap: high & 964 & 87.1 & 77.5 & +9.6 {[}+6.5, +12.8{]} \\
Seed 0, Format: question & 725 & 84.0 & 72.0 & +12.0 {[}+8.1, +15.9{]} \\
Seed 0, Format: number & 935 & 85.7 & 72.4 & +13.3 {[}+9.8, +16.7{]} \\
Seed 0, Format: quote & 325 & 80.6 & 71.1 & +9.5 {[}+3.7, +15.4{]} \\
Seed 0, Format: first person & 687 & 84.9 & 72.1 & +12.8 {[}+8.9, +16.7{]} \\
Llama 3.2 1B, seed 0 & 2{,}892 & 84.9 & 72.1 & +12.8 {[}+10.8, +14.7{]} \\
\bottomrule
\end{tabular}
\end{table}

%% file: tables/e3.tex
\begin{table*}[t]
\centering
\scriptsize
\setlength{\tabcolsep}{3pt}
\caption{Held-out formats: leaving a headline format out of training, and training on old data only. Target pairs: test pairs in which at least one headline has the format (time shift: test pairs created on or after 2015-01-01). Train pairs: training pairs in which no headline has the format (time shift: created before 2013-06-25). M1 in \% with 95\% bootstrap intervals; differences in points (held-out minus full training, same target pairs). $^\dagger$ the 95\% interval excludes 0; $^\ddagger$ also rejected by Holm over the five conditions at $\alpha=0.05$. MDE80: minimum detectable drop at 80\% power from the paired bootstrap SE.\@ LoRA columns: (c) retrained on the same reduced data; drop gap = (c) difference minus (a) difference (positive: (c) loses less). (c) was registered for the held-out-format comparison on 12 Nov 2026 as the then strongest baseline, before (d) was scored.}
\label{tab:e3}
\resizebox{\textwidth}{!}{%
\begin{tabular}{lrrlllrll}
\toprule
 & & & \multicolumn{4}{l}{Classifier (a)} & \multicolumn{2}{l}{LoRA (c)} \\
Held out & Target & Train & Full train & Held-out train & Difference & MDE80 & Difference & Drop gap \\
\midrule
Question mark & 725 & 3{,}288 & 77.9 {[}74.9, 81.0{]} & 77.4 {[}74.3, 80.4{]} & $-$0.6 {[}$-$2.2, +1.1{]} & 2.4 & +0.0 {[}$-$2.2, +2.2{]} & +0.6 {[}$-$2.2, +3.3{]} \\
Digit & 935 & 3{,}639 & 75.5 {[}72.7, 78.2{]} & 73.4 {[}70.6, 76.1{]} & $-$2.1 {[}$-$4.1, $-$0.3{]}$^\dagger$ & 2.7 & $-$2.0 {[}$-$4.4, +0.3{]} & +0.1 {[}$-$2.9, +3.0{]} \\
Quotation & 325 & 4{,}425 & 72.9 {[}68.0, 77.5{]} & 70.2 {[}65.2, 75.1{]} & $-$2.8 {[}$-$5.8, +0.3{]} & 4.2 & +1.2 {[}$-$2.8, +5.2{]} & +4.0 {[}$-$1.2, +9.2{]} \\
First person & 687 & 3{,}906 & 76.0 {[}72.8, 79.2{]} & 76.1 {[}72.9, 79.3{]} & +0.1 {[}$-$1.6, +1.9{]} & 2.4 & $-$1.5 {[}$-$3.9, +0.9{]} & $-$1.6 {[}$-$4.5, +1.3{]} \\
\midrule
Time shift 2013$\to$2015 & 1107 & 882 & 76.3 {[}73.8, 78.8{]} & 73.1 {[}70.4, 75.7{]} & $-$3.3 {[}$-$5.6, $-$0.9{]}$^\ddagger$ & 3.3 & $-$4.1 {[}$-$6.5, $-$1.6{]}$^\ddagger$ & $-$0.8 {[}$-$4.1, +2.4{]} \\
\bottomrule
\end{tabular}}
\end{table*}

%% file: tables/e2.tex
\begin{table*}[t]
\centering
\scriptsize
\setlength{\tabcolsep}{3pt}
\caption{Asked predictions versus revealed clicks (all 2,892 test pairs): correlation of the A$-$B feature difference (presence; word count for length) with the actual winner (first column, 95\% bootstrap interval) or with each method's pick. In parentheses: excess $D$ over a predictor of equal pair accuracy whose errors ignore the features (accuracy re-measured in every resample), $(2\,\mathrm{acc}-1)\times$ actual. $^\dagger$ interval of $D$ excludes 0; $^\ddagger$ also Holm-rejected ($\alpha=0.05$, one family of 63: the eight predictors shown and the count-level model); $^s$ pick correlation's 95\% interval excludes 0 with the sign opposite to actual. ``Asked'': Qwen3-8B-Base without and with adapter (e), asked ``Which headline would you click on?''; Opus: one post-hoc full-set run of (f), not repeated, asked which headline got the higher click-through rate (CTR), not used for the objective comparison. For grounded models $D$ is not evidence of an asked--revealed gap, since their errors depend on the features. $D$ and Holm were added after the correlations were seen. $D$ is a sensitivity reading; the main check of the asked--revealed gap, $G$, is in Appendix~\ref{app:e2cond}; the other readings are in the supplementary archive.}
\label{tab:e2}
\resizebox{\textwidth}{!}{%
\begin{tabular}{llllllllll}
\toprule
 & & Asked as a reader & Asked to predict CTR & \multicolumn{6}{l}{Grounded in clicks} \\
Feature & Actual & Base 8B, asked & Opus 5.5 judge & Fine-tuned 8B, asked & (a) embedding & (b) TF-IDF & (c) LOLA-style & (d) Suh-style & (e) pairwise \\
\midrule
you/your & $-$0.010 {[}$-$0.047, +0.027{]} & +0.018 (+0.019) & $-$0.010 ($-$0.006) & $-$0.017 ($-$0.009) & $-$0.030 ($-$0.025) & $-$0.013 ($-$0.009) & +0.009 (+0.015) & $-$0.029 ($-$0.021) & $-$0.013 ($-$0.006) \\
I/me/my & +0.109 {[}+0.073, +0.145{]} & +0.048 (+0.040$^{\dagger}$) & +0.064 (+0.016) & +0.106 (+0.028$^{\dagger}$) & +0.114 (+0.060$^{\ddagger}$) & +0.132 (+0.090$^{\ddagger}$) & +0.100 (+0.028$^{\dagger}$) & +0.117 (+0.037$^{\dagger}$) & +0.090 (+0.013) \\
question mark & $-$0.174 {[}$-$0.208, $-$0.140{]} & +0.075 (+0.088$^{\ddagger,s}$) & $-$0.072 (+0.005) & $-$0.165 ($-$0.040$^{\ddagger}$) & $-$0.161 ($-$0.075$^{\ddagger}$) & $-$0.142 ($-$0.075$^{\ddagger}$) & $-$0.187 ($-$0.073$^{\ddagger}$) & $-$0.176 ($-$0.048$^{\ddagger}$) & $-$0.153 ($-$0.030$^{\dagger}$) \\
digit & +0.111 {[}+0.074, +0.146{]} & +0.070 (+0.062$^{\ddagger}$) & +0.121 (+0.072$^{\ddagger}$) & +0.117 (+0.037$^{\ddagger}$) & +0.125 (+0.070$^{\ddagger}$) & +0.052 (+0.009) & +0.102 (+0.030$^{\dagger}$) & +0.124 (+0.043$^{\ddagger}$) & +0.117 (+0.038$^{\dagger}$) \\
negation & +0.035 {[}$-$0.002, +0.070{]} & $-$0.054 ($-$0.057$^{\dagger,s}$) & +0.002 ($-$0.014) & +0.044 (+0.019) & +0.063 (+0.046$^{\dagger}$) & +0.063 (+0.049$^{\dagger}$) & +0.038 (+0.015) & +0.057 (+0.032$^{\dagger}$) & +0.047 (+0.023) \\
length (words) & +0.118 {[}+0.082, +0.153{]} & +0.151 (+0.142$^{\ddagger}$) & +0.171 (+0.119$^{\ddagger}$) & +0.160 (+0.076$^{\ddagger}$) & +0.150 (+0.092$^{\ddagger}$) & +0.084 (+0.038$^{\dagger}$) & +0.151 (+0.073$^{\ddagger}$) & +0.160 (+0.074$^{\ddagger}$) & +0.162 (+0.078$^{\ddagger}$) \\
emphasis & $-$0.030 {[}$-$0.066, +0.007{]} & +0.046 (+0.048$^{\dagger,s}$) & $-$0.024 ($-$0.011) & $-$0.045 ($-$0.024$^{\dagger}$) & $-$0.010 (+0.005) & $-$0.028 ($-$0.017) & $-$0.036 ($-$0.016) & $-$0.047 ($-$0.026$^{\dagger}$) & $-$0.028 ($-$0.007) \\
\bottomrule
\end{tabular}}
\end{table*}

%% file: tables/e2full.tex
\begin{table}[ht]
\centering
\scriptsize
\setlength{\tabcolsep}{2.5pt}
\caption{Three models asked two questions on the 2,292 registered test pairs: the 2,892 test pairs minus the 600-pair pilot subset, whose answers had been seen before the rule (rule dated after the objective comparison, which is unchanged). CTR: which headline got the higher click rate; reader: which headline you would click. A pick is the mean over both orders (0, 0.5 or 1). Accuracy gives a 0.5 pick half credit. Counting it as an error (the M1 rule) gives CTR/reader accuracies of 54.0/51.9 (Sonnet 5.5), 44.9/30.1 (Haiku 5.5), 33.2/15.1 (Qwen3-8B), so reader$-$CTR is $-$2.1 to $-$18.1 points under that rule ($-$2.5 to $-$10.3 with half credit); on pairs without a 0.5 pick they are 75.0/71.3, 67.1/51.9, 62.1/56.6 (post hoc; feature correlations on those pairs in \texttt{results/e2\_tie\_rules.json}). Accuracy in \%, order agreement = share of pairs with the same pick in both orders. Feature columns: correlation of the pick with the A$-$B feature difference. $^*$ Holm-rejected ($\alpha=0.05$) asked$-$actual difference within the family of 42 tests. $^\dagger$ Unanswered calls count as a wrong pick in that order (Haiku 5.5 CTR: 1 of 4,584 calls; Haiku 5.5 reader: 37 of 4,584 calls), so in these rows half-credit accuracy is not the M1-rule accuracy plus half the order disagreement, and pairs without a 0.5 pick are not the order-agreeing pairs (Haiku 5.5 CTR: 66.9\% vs.\ 66.8\%; Haiku 5.5 reader: 57.9\% vs.\ 56.3\%). Qwen3-8B is the 4-bit quantized instruct model, not the bf16 original. Experiment-level bootstrap, 10,000 resamples.}
\label{tab:e2full}
\resizebox{\textwidth}{!}{%
\begin{tabular}{llrrrrrrrrr}
\toprule
Model & Question & Accuracy [95\%] & Order agr. & you/your & I/me/my & question mark & digit & negation & length (words) & emphasis \\
\midrule
Actual & & & & $-$0.024 & +0.096 & $-$0.185 & +0.099 & +0.036 & +0.112 & $-$0.051 \\
\midrule
Sonnet 5.5 & CTR & 68.0 {[}66.4, 69.6{]} & 72.0 & $-$0.012 & +0.071 & $-$0.061$^*$ & +0.115 & +0.026 & +0.222$^*$ & $-$0.012 \\
 & reader & 65.5 {[}63.9, 67.1{]} & 72.7 & $-$0.027 & +0.048 & $-$0.030$^*$ & +0.119 & $-$0.014 & +0.178 & $-$0.029 \\
Haiku 5.5 & CTR$^\dagger$ & 61.4 {[}59.8, 63.0{]} & 66.8 & +0.048 & +0.077 & +0.061$^*$ & +0.155 & +0.040 & +0.140 & +0.033$^*$ \\
 & reader$^\dagger$ & 51.1 {[}49.6, 52.6{]} & 56.3 & $-$0.095 & $-$0.010$^*$ & +0.022$^*$ & +0.057 & $-$0.031 & +0.057 & +0.006 \\
Qwen3-8B, 4-bit & CTR & 56.5 {[}55.0, 58.0{]} & 53.4 & +0.018 & +0.063 & +0.017$^*$ & +0.126 & $-$0.029 & $-$0.004$^*$ & +0.062$^*$ \\
 & reader & 51.8 {[}50.7, 52.9{]} & 26.7 & $-$0.045 & +0.041 & $-$0.029$^*$ & +0.051 & $-$0.017 & +0.082 & +0.026 \\
\bottomrule
\end{tabular}}
\end{table}

%% file: tables/behaviors.tex
\begin{table}[ht]
\centering
\caption{Behaviors and datasets considered in this paper. ``Status'' is read from the committed result files: \emph{main text} = reported in Section~\ref{sec:res-obj}, \emph{appendix} = reported in the appendix, \emph{not run} = no result file. Exposure tells whether which item a user saw was randomized.}
\label{tab:behaviors}
\scriptsize
\setlength{\tabcolsep}{3pt}
\begin{tabular}{@{}l p{2.2cm} p{1.6cm} p{1.4cm} p{3.4cm} r p{1.4cm}@{}}
\toprule
Behavior & Data & Unit & Exposure & Pair definition & Test pairs & Status \\
\midrule
Click & {\raggedright Upworthy A/B archive\par} & {\raggedright headline (same story)\par} & {\raggedright random (A/B test)\par} & {\raggedright two headlines of one test whose click rates differ significantly\par} & 2,892 & {\raggedright main text\par} \\
Click & {\raggedright MIND-small (MSN logs; research-use license, not redistributed)\par} & {\raggedright news title (one user)\par} & {\raggedright recommender\par} & {\raggedright main: the clicked title among one user's impression list (1 clicked, up to 4 not), with and without the user's last 10 clicked titles; auxiliary: two articles of one list by aggregate click rate, not run: 821 test pairs remain after removing articles seen in training, below the pre-registered 1,000\par} & -- & {\raggedright appendix\par} \\
Watch time & {\raggedright KuaiRand-Pure (valid play)\par} & {\raggedright video caption\par} & {\raggedright random insertion\par} & {\raggedright same user and day, one video with \texttt{long\_view}=1 and one with 0, same length bin\par} & 3,000 & {\raggedright appendix\par} \\
Likes per view & {\raggedright MicroLens-100K (platform totals; academic use, videos not redistributed)\par} & {\raggedright video title and category; + descriptions of 4 frames\par} & {\raggedright recommender\par} & {\raggedright one video, top versus bottom quartile of likes per view, balanced by category and first-interaction quartile; 80/20 random video split\par} & 942 videos & {\raggedright appendix\par} \\
\bottomrule
\end{tabular}
\end{table}

%% file: tables/ext_full.tex
\begin{table}[t]
\centering
\scriptsize
\setlength{\tabcolsep}{3pt}
\caption{Asked versus grounded beyond headlines: pair accuracy (\%) with 95\% intervals (small print) from a user-level bootstrap (10,000 resamples). KuaiRand pairs are one long-viewed and one not long-viewed video of the same user, day and length bin under random exposure; the subset rows use the 1,000 pairs that were also asked to the commercial judges through the API (OpenRouter), each pair in both orders. ``Past rate'' is the video's long-view rate in the earlier standard-feed log; ``+history'' adds the viewer's recent watch history to the prompt or input. -- = not part of the design for that row. Differences inside this table were not tested unless reported in the text. MIND rows are top-1 hit rate, not pair accuracy: the share of test impressions whose clicked article the method ranks first among the shown candidates (uniform guessing 21.8\%, 22.1\% on the subset), user-level bootstrap with 2,000 resamples; ``Past rate'' is the article's click-through rate in the training week; the confirmation sample is 3,000 further test impressions of other users, drawn and registered before asking, and is not pooled with the subset; on the 1,000 registered impressions the same comparison (grounded 8B with history minus asked Sonnet as this reader) is +2.1 top-1 points [$-$0.8, +5.1]; every grounded 8B here, including the two behind the history effect, is a single training run (seed 0); the asked judges are prompted as this reader, each impression in both candidate orders, through the Claude Code command-line client, and an unanswered order counts as a uniform guess. MicroLens rows are single-video accuracy (top versus bottom quartile of likes per view, half each, video-level bootstrap with 2,000 resamples), from title and category only; for MicroLens the ``Past rate'' column is video length alone (a logistic regression on log length); the MicroLens asked cells are Sonnet through the API (temperature 0, no added context), as registered for the path check after the first answers had come through the Claude Code command-line client (default system prompt and sampling); with those earlier answers asked Sonnet scores 57.1\% on title (54.6\% length-matched). The ``+ frames'' row adds a written description of the video's frames to the title and category; its third line ($\Delta$) is the change from title only in points on the same videos (paired video bootstrap, 2,000 resamples, Holm over the three methods): embedding +7.0 [+4.6, +9.7]; grounded 8B +7.3 [+4.7, +10.1]; asked Sonnet +1.1 [$-$2.5, +4.7]. The rise is significant after Holm for embedding and grounded 8B; asked Sonnet does not change. The ``new videos'' rows are 200 further MicroLens videos outside the first sample (category-matched, half top quartile), drawn, registered and asked before the already grounded 8B models scored them, and not pooled with the first sample. Registered test, 8B minus asked Sonnet with Holm over the two versions: title +11.0 [+1.5, +20.5], + frames +20.0 [+11.0, +29.0]; both rejected, a pre-registered confirmation (on the earlier command-line answers; with the API answers in the rows, title +9.0 [+0.0, +18.5] with Holm $p=0.067$ and + frames +20.0 [+11.5, +28.5] with Holm $p<0.001$, + frames rejected only). On the 124 length-matched new videos the differences (command-line answers) are title +8.1 [$-$4.8, +21.0] and + frames +15.3 [+3.2, +26.6] (the title interval includes zero). Exposure is removed by design in the KuaiRand pairs, yet every method that ignores the viewer stays near chance, so these text representations cannot decide whether asking errs where training succeeds. The +history column of the grounded model reflects a training setup that did not learn to use the history input, not evidence that history is useless.}
\label{tab:extfull}
\begin{tabular}{@{}l r cc cc cc@{}}
\toprule
 & & & & \multicolumn{2}{c}{Asked LLM} & \multicolumn{2}{c}{Grounded 8B} \\
\cmidrule(lr){5-6}\cmidrule(lr){7-8}
Data (behavior) & Pairs & Past rate & Embedding & plain & +history & plain & +history \\
\midrule
KuaiRand-Pure (watch time) & 3,000 & 53.6 & 55.2 & -- & -- & 56.2 & 56.8 \\
 & & {\tiny [51.8, 55.4]} & {\tiny [53.4, 57.0]} &  &  & {\tiny [54.5, 58.0]} & {\tiny [55.0, 58.6]} \\
\quad subset, asked Sonnet & 1,000 & 53.4 & 56.1 & 53.5 & 59.4 & 55.9 & 56.1 \\
 & & {\tiny [50.3, 56.5]} & {\tiny [53.0, 59.2]} & {\tiny [50.7, 56.2]} & {\tiny [56.7, 62.0]} & {\tiny [52.7, 59.0]} & {\tiny [53.0, 59.2]} \\
\quad same subset, asked Haiku &  &  &  & 52.2 & 54.1 &  &  \\
 & &  &  & {\tiny [49.5, 54.9]} & {\tiny [51.8, 56.5]} &  &  \\
MIND-small (click), impressions & 20,000 & 26.1 & -- & -- & -- & 28.9 & 38.2 \\
 & & {\tiny [25.5, 26.7]} &  &  &  & {\tiny [28.3, 29.6]} & {\tiny [37.5, 38.9]} \\
\quad subset, asked Sonnet & 1,000 & 26.1 & -- & 28.7 & 38.1 & 31.0 & 40.2 \\
 & & {\tiny [23.6, 28.6]} &  & {\tiny [26.4, 31.2]} & {\tiny [35.3, 40.8]} & {\tiny [28.1, 33.9]} & {\tiny [37.3, 43.2]} \\
\quad same subset, asked Haiku &  &  &  & 24.2 & 31.4 &  &  \\
 & &  &  & {\tiny [22.1, 26.5]} & {\tiny [29.2, 33.3]} &  &  \\
\quad confirmation sample & 3,000 & -- & -- & -- & 38.6 & -- & 38.1 \\
 & &  &  &  & {\tiny [36.8, 40.3]} &  & {\tiny [36.4, 39.9]} \\
MicroLens-100K (likes/view), videos & 942 & 68.6 & 66.9 & 60.6 & -- & 71.8 & -- \\
 & & {\tiny [65.6, 71.5]} & {\tiny [63.7, 69.8]} & {\tiny [57.4, 63.8]} &  & {\tiny [69.1, 74.5]} &  \\
\quad + frames & 942 & -- & 73.9 & 61.7 & -- & 79.1 & -- \\
 & &  & {\tiny [71.1, 76.5]} & {\tiny [58.5, 64.6]} &  & {\tiny [76.6, 81.5]} &  \\
 & &  & {\tiny $\Delta$+7.0} & {\tiny $\Delta$+1.1} &  & {\tiny $\Delta$+7.3} &  \\
\quad new videos, title & 200 & -- & -- & 59.0 & -- & 68.0 & -- \\
 & &  &  & {\tiny [52.0, 65.5]} &  & {\tiny [61.5, 74.0]} &  \\
\quad new videos, + frames & 200 & -- & -- & 58.0 & -- & 78.0 & -- \\
 & &  &  & {\tiny [51.0, 64.5]} &  & {\tiny [72.0, 83.5]} &  \\
\bottomrule
\end{tabular}
\end{table}

%% file: tables/seeds.tex
\begin{table}[ht]
\centering
\scriptsize
\setlength{\tabcolsep}{3pt}
\caption{Pair accuracy M1 (\%) on the 2,892 test pairs for each training seed (auxiliary; the main text reports seed 0). Mean: per-pair correctness averaged over the three seeds, with a 95\% bootstrap interval over experiments. Seed 0 of the count-level model, the one the main text reports, is the highest of its three seeds; on the three-seed means it differs from (d) by +0.40 points [$-$0.45, +1.27]. On the 600-pair subset the order of the three-seed means is reversed (86.17 against 86.44 for (d)). Three seeds give only a coarse estimate of how much results vary between seeds, so intervals that resample seeds are themselves rough.}
\label{tab:seeds}
\begin{tabular}{llllll}
\toprule
Method & Seed 0 & Seed 1 & Seed 2 & Mean [95\% CI] & Mean (600) \\
\midrule
Count-level objective (ours) & 87.45 & 86.76 & 87.03 & 87.08 {[}85.95, 88.17{]} & 86.17 \\
(d) Suh-style option KL & 86.55 & 86.58 & 86.89 & 86.68 {[}85.52, 87.81{]} & 86.44 \\
(e) Pairwise LoRA classifier & 85.48 & 85.55 & 84.92 & 85.32 {[}84.15, 86.45{]} & 85.94 \\
(c) LOLA-style CTR regression & 82.75 & 82.57 & 83.47 & 82.93 {[}81.74, 84.08{]} & 82.61 \\
\bottomrule
\end{tabular}
\end{table}

%% file: tables/prompts.tex
\paragraph{Judge prompt (f): system message of the commercial judges.}\mbox{}\par\nobreak
\begin{lstlisting}[breaklines=true,basicstyle=\scriptsize\ttfamily,columns=fullflexible,upquote=true,frame=single]
You are an expert in online media. Upworthy ran A/B tests in 2014-2015 showing different headlines for the same story (same image) to Facebook users. For each numbered pair, say which headline (X or Y) got the higher click-through rate. Answer one line per pair, like '1: X'. No explanations.
\end{lstlisting}

\paragraph{Pair line (f, g): one line per pair in the user message of the judges and of the personas.}\mbox{}\par\nobreak
\begin{lstlisting}[breaklines=true,basicstyle=\scriptsize\ttfamily,columns=fullflexible,upquote=true,frame=single]
{i}. Post X: "{first}"
   Post Y: "{second}"
\end{lstlisting}

\paragraph{MoP persona-writing prompt (g): system message that writes each persona.}\mbox{}\par\nobreak
\begin{lstlisting}[breaklines=true,basicstyle=\scriptsize\ttfamily,columns=fullflexible,upquote=true,frame=single]
You study audiences of a viral news site. Upworthy ran headline A/B tests on Facebook in 2013-2014. Below are headline pairs for the same story; the first headline of each pair got the higher click-through rate. Describe, in 2-3 sentences written in the second person ("You are ..."), the kind of Facebook reader whose clicking produced these winners: who they are, what draws them in, and what turns them off. Output only the description.
\end{lstlisting}

\paragraph{MoP exemplar line (g): one line per training pair, used to write personas and as in-context examples.}\mbox{}\par\nobreak
\begin{lstlisting}[breaklines=true,basicstyle=\scriptsize\ttfamily,columns=fullflexible,upquote=true,frame=single]
{i}. Clicked more: "{win}"
   Clicked less: "{lose}"
\end{lstlisting}

\paragraph{MoP persona prompt (g): system message of each persona.}\mbox{}\par\nobreak
\begin{lstlisting}[breaklines=true,basicstyle=\scriptsize\ttfamily,columns=fullflexible,upquote=true,frame=single]
{persona description} It is late 2014 and you are scrolling your Facebook news feed.

Here are earlier pairs where you (and readers like you) clicked one headline more than the other:
{exemplar pairs}

Now pairs of posts appear; in each pair both posts share the same picture and only the headline differs. For each numbered pair, decide as yourself which post you would click, X or Y. Answer one line per pair, like '1: X'. No explanations.
\end{lstlisting}

\paragraph{Asked 8B prompt: the reader question put to the base and the fine-tuned 8B model in the feature diagnostic.}\mbox{}\par\nobreak
\begin{lstlisting}[breaklines=true,basicstyle=\scriptsize\ttfamily,columns=fullflexible,upquote=true,frame=single]
You are scrolling through your social media feed. Two posts link to the same article but use different headlines.
A: {a}
B: {b}
Which headline would you click on?
Answer:
\end{lstlisting}

%% file: tables/extrain.tex
\section{Training runs and prompts of the extension datasets}
\label{app:exttrain}
Every grounded 8B arm of Table~\ref{tab:extfull} uses the settings of (e) without reselection: peak learning rate $3\times10^{-5}$ (the rate (c), (d) and (e) each chose on Upworthy), 20 warm-up steps, clipping 1, 2 epochs, batch 16, LoRA rank 16 and scale 2, seed 0, the Qwen3-8B-Base model of Appendix~\ref{app:training} and 43.6M trainable parameters. No development score was used to pick anything; development items were only watched. Each table (Tables~\ref{tab:trainkw},~\ref{tab:trainmind} and~\ref{tab:trainml} for KuaiRand, MIND and MicroLens) lists the runs as logged (Updates = epochs $\times$ examples / batch, rounded up per epoch; minutes include periodic development scoring, Total until scoring finished). Prompts follow in the format of Appendix~\ref{app:prompts}; braces are filled per item.

\begin{table}[ht]
\centering
\scriptsize
\caption{KuaiRand watch time, grounded 8B: one example is a pair of videos shown to the same user on the same day, one of them long-viewed; first token `` A'' vs.\ `` B'' after ``Answer:'', cross-entropy on the long-viewed one, each pair in both orders (as (e)). 20,000 training pairs from the standard feed ($2\times$ = examples), 1,000 development pairs watched only. The + user and history arm adds the user's 5 profile fields (user\_active\_degree, follow\_user\_num\_range, fans\_user\_num\_range, friend\_user\_num\_range, register\_days\_range) and up to 10 captions the user last long-viewed, oldest first; captions are cut to 96 tokens and, past 768 tokens, the oldest history captions are dropped first. Scored: development and three test splits.}
\label{tab:trainkw}
\begin{tabular}{lrrrrrrr}
\toprule
Arm & Examples & Monitor & Micro-batch & Updates & Train min & Total min & Scored \\
\midrule
Captions only & 40,000 & 1,000 & -- & 5,000 & 193 & 212 & 10,000 \\
+ user and history & 40,000 & 1,000 & 2 & 5,000 & 586 & 653 & 10,000 \\
\bottomrule
\end{tabular}
\end{table}

\paragraph{Prompts, KuaiRand watch time.}
Grounded 8B input; the + user and history arm puts the first two lines and the history list before the question, the other arm starts at the question:
\begin{lstlisting}[breaklines=true,basicstyle=\scriptsize\ttfamily,columns=fullflexible,upquote=true,frame=single]
User: {user features}
Recently watched to the end or past 18 seconds:
- {oldest caption}
- ...
- {latest caption}
Two short videos were shown to the same user on the same day. Which one did the user watch for longer than 18 seconds or to the end?
A: {caption of video A}
B: {caption of video B}
Answer:
\end{lstlisting}
Asked system message with user and history:
\begin{lstlisting}[breaklines=true,basicstyle=\scriptsize\ttfamily,columns=fullflexible,upquote=true,frame=single]
You are an expert on short-video apps. Kuaishou, a Chinese short-video app, showed one user two videos on the same day in its single-column feed. For each video you see its caption (in Chinese). You also see this user's profile and the captions of the videos this user most recently watched for a long view, oldest first. The user watched exactly one of the two for a long view: more than about 18 seconds, or to the end. Which video (X or Y) did this user watch for a long view? Answer one line, like '1: X'. No explanations.
\end{lstlisting}
Asked system message, captions only:
\begin{lstlisting}[breaklines=true,basicstyle=\scriptsize\ttfamily,columns=fullflexible,upquote=true,frame=single]
You are an expert on short-video apps. Kuaishou, a Chinese short-video app, showed one user two videos on the same day in its single-column feed. For each video you see its caption (in Chinese). The user watched exactly one of the two for a long view: more than about 18 seconds, or to the end. Which video (X or Y) did this user watch for a long view? Answer one line, like '1: X'. No explanations.
\end{lstlisting}
Asked user message with user and history (captions only: the last two lines):
\begin{lstlisting}[breaklines=true,basicstyle=\scriptsize\ttfamily,columns=fullflexible,upquote=true,frame=single]
User profile: {user features}
Recently long-viewed videos (oldest first):
- "{oldest caption}"
- "..."
- "{latest caption}"

1. Video X: "{caption X}"
   Video Y: "{caption Y}"
\end{lstlisting}

\begin{table}[ht]
\centering
\scriptsize
\caption{MIND clicks, grounded 8B: one example is an impression list (the clicked headline and up to four unclicked ones, letters A--E); softmax over the shown letters, cross-entropy on the clicked one, given and reversed order. 20,000 training impressions ($2\times$ = examples), 1,000 development impressions watched only. The + clicked history arm adds up to 10 headlines the reader clicked before, oldest first, with input capped at 512 tokens (longest real input below it). Scored: development and test impressions.}
\label{tab:trainmind}
\begin{tabular}{lrrrrrrr}
\toprule
Arm & Examples & Monitor & Micro-batch & Updates & Train min & Total min & Scored \\
\midrule
Headlines only & 40,000 & 1,000 & -- & 5,000 & 164 & 198 & 21,000 \\
+ clicked history & 40,000 & 1,000 & 2 & 5,000 & 486 & 589 & 21,000 \\
\bottomrule
\end{tabular}
\end{table}

\paragraph{Prompts, MIND news clicks.}
Grounded 8B input with clicked history (three candidates shown, up to five); the headlines-only arm starts at the first line after the history:
\begin{lstlisting}[breaklines=true,basicstyle=\scriptsize\ttfamily,columns=fullflexible,upquote=true,frame=single]
Headlines this reader clicked before (oldest first):
- {oldest title}
- ...
- {latest title}
News headlines shown to a reader on a news site:
A: {title 1}
B: {title 2}
C: {title 3}
Which one did the reader click?
Answer:
\end{lstlisting}
Asked system message, reader question with history; without history the sentence starting ``You also see'' is left out:
\begin{lstlisting}[breaklines=true,basicstyle=\scriptsize\ttfamily,columns=fullflexible,upquote=true,frame=single]
You are an expert on online news. Microsoft News (MSN) showed one user a personalized list of news articles in late 2019. You see the headlines of the articles in that list, labelled with letters. You also see the headlines of the articles this user most recently clicked before, oldest first. The user clicked exactly one of the listed articles. Which article did this user click? Answer one line with its letter, like '1: B'. No explanations.
\end{lstlisting}
Question that replaces the reader question in the CTR arms:
\begin{lstlisting}[breaklines=true,basicstyle=\scriptsize\ttfamily,columns=fullflexible,upquote=true,frame=single]
Across all users who saw these articles that week, which listed article got the highest click-through rate? Answer one line with its letter, like '1: B'. No explanations.
\end{lstlisting}
Asked user message with history (headlines only: the article list alone):
\begin{lstlisting}[breaklines=true,basicstyle=\scriptsize\ttfamily,columns=fullflexible,upquote=true,frame=single]
Recently clicked headlines (oldest first):
- "{oldest title}"
- "..."
- "{latest title}"

1. Articles:
   A. "{title 1}"
   B. "{title 2}"
   C. "{title 3}"
\end{lstlisting}

\begin{table}[ht]
\centering
\scriptsize
\caption{MicroLens likes per view, grounded 8B: one example is one video; first token `` high'' vs.\ `` low'' after ``Answer:'', cross-entropy on its class (top or bottom quartile of likes per view). Examples = training videos of each sample minus a 10\% development split watched only (Monitor); the length-matched sample is the auxiliary one of Table~\ref{tab:extfull} and is trained on its own training videos. Titles cut to 96 tokens. Frame runs accumulate the batch of 16 over two parts of 8 (memory). Scored: development and test videos.}
\label{tab:trainml}
\begin{tabular}{lrrrrrrr}
\toprule
Arm & Examples & Monitor & Micro-batch & Updates & Train min & Total min & Scored \\
\midrule
Title, category; first sample & 3,425 & 380 & -- & 430 & 19 & 19 & 1,322 \\
Title, category; length-matched & 2,260 & 251 & -- & 284 & 6 & 7 & 872 \\
+ frames; first sample & 3,425 & 380 & 8 & 430 & 28 & 30 & 1,322 \\
+ frames; length-matched & 2,260 & 251 & 8 & 284 & 18 & 19 & 872 \\
\bottomrule
\end{tabular}
\end{table}

\paragraph{Prompts, MicroLens likes per view.}
Grounded 8B input with frames; title-and-category runs leave out the Frames line:
\begin{lstlisting}[breaklines=true,basicstyle=\scriptsize\ttfamily,columns=fullflexible,upquote=true,frame=single]
A short video on a video-sharing platform:
Title: "{title}"
Category: {category}
Frames: "{frame description}"
Is its like-to-view ratio in the top 25% (high) or the bottom 25% (low) of the platform's videos?
Answer:
\end{lstlisting}
Asked system message, title and category; the user message is the Title and Category lines of the training input:
\begin{lstlisting}[breaklines=true,basicstyle=\scriptsize\ttfamily,columns=fullflexible,upquote=true,frame=single]
You are an expert on short-video platforms. MicroLens is a Chinese short-video platform. The videos you will be asked about were sorted by their ratio of likes to views on the platform: half of them are in the top 25% of all videos by this ratio (high), and half are in the bottom 25% (low). You see one video's title (translated to English) and its category. Is this video's likes-per-view ratio high (top 25%) or low (bottom 25%)? Answer with one word: high or low. No explanations.
\end{lstlisting}
Asked system message with frames; the user message adds the Frames line:
\begin{lstlisting}[breaklines=true,basicstyle=\scriptsize\ttfamily,columns=fullflexible,upquote=true,frame=single]
You are an expert on short-video platforms. MicroLens is a Chinese short-video platform. The videos you will be asked about were sorted by their ratio of likes to views on the platform: half of them are in the top 25% of all videos by this ratio (high), and half are in the bottom 25% (low). You see one video's title (translated to English), its category, and a neutral description of four frames taken from the video. Is this video's likes-per-view ratio high (top 25%) or low (bottom 25%)? Answer with one word: high or low. No explanations.
\end{lstlisting}

%% file: tables/repro.tex
\begin{table}[ht]
\centering
\scriptsize
\setlength{\tabcolsep}{3pt}
\caption{Final LoRA runs: prompt string (\texttt{\textbackslash n} is a newline, braces are filled per example), units or pairs per batch, epochs over the full training set, total optimizer updates, peak learning rate (20 warm-up steps, gradient clipping 1.0), LoRA rank and scale. Base model and seed 0 are shared. At one learning rate (d) takes twice the updates of the count-level objective (1,884 vs.\ 942) because its batch is half as large. (d) chose its learning rate at batch 16; its final run was restarted at batch 8 from scratch when a speed watchdog stopped the batch-16 run at step 280 (tuning budget in the supplementary appendix to be released with the code{}).}
\label{tab:repro}
\begin{tabular}{p{0.19\linewidth}p{0.38\linewidth}rrrrr}
\toprule
Method & Prompt & Batch & Epochs & Updates & LR & Rank/scale \\
\midrule
(c) LOLA-style regression & \texttt{Headline: \{h\}\textbackslash{}nHeadline:} & 16 & 2 & 3,916 & $3\times10^{-5}$ & 16/2 \\
(d) Suh-style KL & \texttt{Several headlines were tested on the same article. Which one did readers click the most?\textbackslash{}nA: \{headline 1\}\textbackslash{}nB: \{headline 2\}\textbackslash{}n...\textbackslash{}nAnswer:} & 8 & 2 & 1,884 & $3\times10^{-5}$ & 16/2 \\
(e) Pair classifier & \texttt{Two headlines were tested on the same article. Which one got the higher click rate?\textbackslash{}nA: \{a\}\textbackslash{}nB: \{b\}\textbackslash{}nAnswer:} & 16 & 2 & 4,242 & $3\times10^{-5}$ & 16/2 \\
Count-level objective & \texttt{Headline: \{h\}\textbackslash{}nWould this reader click this headline? Answer click or skip.\textbackslash{}nAnswer:} & 16 & 2 & 942 & $3\times10^{-5}$ & 16/2 \\
\bottomrule
\end{tabular}
\end{table}

\begin{table}[ht]
\centering
\scriptsize
\caption{Final LoRA runs, as logged: training units, i.e.\ experiments (all units of the training period, none held out), fitted examples (headlines for (c), experiments with at least two headlines and one click for (d) and ours, ordered significant pairs for (e)), trainable parameters, minutes of training including periodic development evaluation, minutes until scoring finished, and experiments scored. One Apple M5 Max, one run at a time, no resumes.}
\label{tab:reprorun}
\begin{tabular}{lrrrrrr}
\toprule
Method & Units & Examples & Trainable & Train min & Total min & Scored \\
\midrule
(c) LOLA-style regression & 7,537 & 31,325 & 43.7M & 43 & 62 & 11,944 \\
(d) Suh-style KL & 7,537 & 7,535 & 43.6M & 43 & 100 & 11,944 \\
(e) Pair classifier & 7,537 & 33,934 & 43.6M & 79 & 160 & 11,944 \\
Count-level objective & 7,537 & 7,535 & 43.7M & 78 & 86 & 4,407 \\
\bottomrule
\end{tabular}
\end{table}

%% file: tables/amend.tex
\providecommand{\AmendN}{11}
\providecommand{\AmendCand}{3}
\providecommand{\AmendModel}{3}
\providecommand{\AmendOther}{5}
\providecommand{\AmendCandGap}{11.9}
\begin{table}[t]
\centering
\small
\caption{The \AmendN{} rule changes committed after baseline results were seen and before any count-level test score, in commit order (times KST, 2026). Each was committed before the run it governs. ``Seen'': baseline results available at the time. ``Bears on'': a candidate for the strongest baseline, the count-level model of the registered comparison, or neither. No change alters the metrics, the split or the rule that picks the strongest baseline.}
\label{tab:amend}
\begin{tabular}{@{}l p{5.8cm} p{2.6cm} p{2.4cm}@{}}
\toprule
Date & Change & Seen & Bears on \\
\midrule
Nov 12, 02:26 & {\raggedright Design of (g): persona clusters, examples, fitted mixture; representative model chosen by training-pair CV\par} & {\raggedright (a)\par} & {\raggedright candidate (g)\par} \\
Nov 12, 02:33 & {\raggedright Calls and scoring of (f) Opus 5.5; unanswered pairs count as wrong; Opus fixed as representative\par} & {\raggedright (a), (g)\par} & {\raggedright candidate (f)\par} \\
Nov 12, 02:37 & {\raggedright C of (a) chosen by 5-fold CV repeated over 5 seeds, because one unshuffled fold order moved C\par} & {\raggedright (a), (g), (f)\par} & {\raggedright candidate (a)\par} \\
Nov 12, 08:06 & {\raggedright Learning-rate multiplier for prompt vectors (1, 30, 300), picked on development runs\par} & {\raggedright (a), (g), (f), (c), feature diagnostic, held-out-format classifier, (e)\par} & {\raggedright count-level training\par} \\
Nov 12, 09:37 & {\raggedright Implementation of the two registered ablations (win/loss only; one shared $\alpha$)\par} & {\raggedright (a), (g), (f), (c), feature diagnostic, held-out-format classifier, (e)\par} & {\raggedright ablations\par} \\
Nov 12, 12:13 & {\raggedright Interpretability test for an extension not reported here (claim later withdrawn)\par} & {\raggedright (a), (g), (f), (c), feature diagnostic, held-out-format classifier, (e), (d)\par} & {\raggedright unreported extension\par} \\
Nov 12, 13:53 & {\raggedright How the number of mixed adapters is chosen on development units (it chose one: the count-level model)\par} & {\raggedright (a), (g), (f), (c), feature diagnostic, held-out-format classifier, (e), (d)\par} & {\raggedright count-level selection\par} \\
Nov 12, 14:25 & {\raggedright Design of the KuaiRand logging-policy analysis, before any estimator run\par} & {\raggedright (a), (g), (f), (c), feature diagnostic, held-out-format classifier, (e), (d)\par} & {\raggedright KuaiRand\par} \\
Nov 13, 00:55 & {\raggedright Which test score enters the comparison: the final retrain of the chosen setting, not a development run\par} & {\raggedright (a), (g), (f), (c), feature diagnostic, held-out-format classifier, (e), (d), KuaiRand\par} & {\raggedright comparison input\par} \\
Nov 13, 02:19 & {\raggedright Order of later GPU runs and what a negative objective-comparison result changes; every registered run reported either way\par} & {\raggedright (a), (g), (f), (c), feature diagnostic, held-out-format classifier, (e), (d), KuaiRand\par} & {\raggedright run order\par} \\
Nov 13, 08:14 & {\raggedright Seeds 1 and 2 of (c), (d), (e) and the count-level model, same settings, auxiliary to the verdict\par} & {\raggedright (a), (g), (f), (c), feature diagnostic, held-out-format classifier, (e), (d), KuaiRand\par} & {\raggedright seeds (auxiliary)\par} \\
\bottomrule
\end{tabular}
\end{table}